\documentclass{ieeeaccess}
\ifCLASSINFOpdf   \else   \fi
\usepackage{cite}
\usepackage{amsmath,amssymb,amsfonts}
\usepackage{algorithmic}
\usepackage{graphicx}
\usepackage{textcomp}
\usepackage{makecell}
\usepackage{array}
\usepackage{booktabs}
\usepackage{soul}
\usepackage{multirow}
\usepackage{float} % For [H] placement specifier
\usepackage{enumitem}
\usepackage{tabularx}
\usepackage{url}
\usepackage[breaklinks]{hyperref}
\usepackage{algorithm}
\usepackage[caption=false,font=footnotesize]{subfig}
\usepackage{soul}
\sethlcolor{yellow}
\renewcommand{\arraystretch}{1.5}    

\usepackage{bm}
\makeatletter
\AtBeginDocument{\DeclareMathVersion{bold}
\SetSymbolFont{operators}{bold}{T1}{times}{b}{n}
\SetSymbolFont{NewLetters}{bold}{T1}{times}{b}{it}
\SetMathAlphabet{\mathrm}{bold}{T1}{times}{b}{n}
\SetMathAlphabet{\mathit}{bold}{T1}{times}{b}{it}
\SetMathAlphabet{\mathbf}{bold}{T1}{times}{b}{n}
\SetMathAlphabet{\mathtt}{bold}{OT1}{pcr}{b}{n}
\SetSymbolFont{symbols}{bold}{OMS}{cmsy}{b}{n}
\renewcommand\boldmath{\@nomath\boldmath\mathversion{bold}}}
\makeatother

\begin{document}
\history{Date of publication xxxx 00, 0000, date of current version xxxx 00, 0000.}
\doi{10.1109/ACCESS.2025.0429000}

\title{D2H-AD: A Hybrid Model Utilizing Hyperdimensional Computing for Advanced Anomaly Detection}

\author{\uppercase{Ghazal~Ghajari}\authorrefmark{1}\href{https://orcid.org/0009-0004-8370-9417}{\includegraphics[height=7pt]{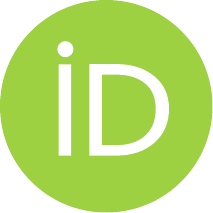}}, \uppercase{Elaheh~Ghajari}\authorrefmark{2}\href{https://orcid.org/0009-0009-4263-5112}{\includegraphics[height=7pt]{logo-orcid.pdf}},
\uppercase{Ashutosh~Ghimire}\authorrefmark{1}\href{https://orcid.org/0000-0001-6210-1219}{\includegraphics[height=7pt]{logo-orcid.pdf}},
\uppercase{Saeid~Ataei}\authorrefmark{3}\href{https://orcid.org/0009-0004-0326-2249}{\includegraphics[height=7pt]{logo-orcid.pdf}},\\
\uppercase{Faris Alsulami}\authorrefmark{4}\href{https://orcid.org/0000-0002-0868-3905}{\includegraphics[height=7pt]{logo-orcid.pdf}},
\uppercase{Fathi~Amsaad}\authorrefmark{1}\href{https://orcid.org/0000-0002-7582-8326}{\includegraphics[height=7pt]{logo-orcid.pdf}}}

\address[1]{Wright State University, Dayton, Ohio 45435, USA (e-mail: [ghajari.2, ashutosh.ghimire, fathi.amsaad]@wright.edu)}
% \address[1]{Wright State University, USA (e-mail: ghajari.2@wright.edu)}
\address[2]{Azad University, Ahvaz 6887561349, Iran (e-mail: elaheh.ghajari.1@gmail.com)}
%\address[3]{Wright State University, USA (e-mail: ashutosh.ghimire@wright.edu)}
\address[3]{Stevens Institute of Technology, Hoboken, New Jersey 07030, USA (e-mail: sataei@stevens.edu)}
\address[4]{University of Jeddah, Jeddah 23890, Saudi Arabia (e-mail: fnalsulami@uj.edu.sa)}

\markboth
{Author \headeretal: Preparation of Papers for IEEE ACCESS}
{Author \headeretal: Preparation of Papers for IEEE TRANSACTIONS and JOURNALS}

%\corresp{Corresponding author: Ghazal~Ghajari (e-mail: ghajari.2@wright.edu).}

%=======================================================================
% ISSUE 7 FIX (Abstract): Clarified that 5.4% gain is over the
% Euclidean-space ablation baseline, not over the five published baselines.
% Added explicit statement about outperforming all five published baselines.
%=======================================================================
\begin{abstract}
Anomaly detection is a cornerstone of modern intelligent systems, with critical applications in healthcare, cybersecurity, smart grids, and IoT environments. While traditional machine learning and deep learning models have shown promise in identifying outliers, they often face challenges such as dependence on large labeled datasets, high computational costs, and limited scalability to edge devices or high-dimensional data streams. This study introduces D2H-AD, a novel anomaly detection framework built upon Hyperdimensional Computing (HDC), a brain-inspired paradigm that encodes information using high-dimensional distributed vectors. Unlike prior HDC-based approaches, D2H-AD fuses distance-based similarity and density-aware encoding in a hybrid design, significantly improving anomaly characterization and detection accuracy. Ablation experiments confirm that hyperdimensional encoding alone contributes up to 5.4\% higher ROC-AUC compared to applying the same density-distance scoring in the original Euclidean feature space, and D2H-AD consistently outperforms all five published baselines (HDAD, ODHD, One-Class SVM, Isolation Forest, and Autoencoders) across every evaluated dataset. The method is designed to be lightweight, interpretable, and computationally efficient, suggesting strong suitability for deployment in resource-constrained and real-time environments. To validate its effectiveness, we evaluate D2H-AD on five diverse benchmark datasets, comparing its performance against five leading techniques: HDAD, ODHD, One-Class SVM, Isolation Forest, and Autoencoders. Our results show that D2H-AD consistently achieves superior F1 scores and ROC-AUC metrics, demonstrating robustness against class imbalance, noise, and data complexity. Beyond accuracy, D2H-AD offers practical advantages including scalability, minimal memory footprint, and an expected low-latency profile derived from its binary operations and lightweight design. These characteristics are crucial for TinyML and edge AI applications. This work highlights the untapped potential of HDC for high-performance anomaly detection and opens new avenues for secure, interpretable, and energy-efficient AI solutions in dynamic environments such as IoT, embedded systems, and beyond. Additionally, D2H-AD provides feature-level interpretability through hypervector decoding, enabling transparent explanations for safety-critical applications.

\end{abstract}


\titlepgskip=-21pt

\maketitle
\section{Introduction}
\label{sec:introduction}
\PARstart{A}{nomaly} detection, also known as outlier detection, is a critical process in data analysis and machine learning that involves identifying unusual patterns or behaviors in data streams that deviate significantly from what is expected. These anomalies or outliers are often sudden, infrequent phenomena that were not previously encountered, making them difficult to predict and manage \cite{agrawal2015survey}. While some anomalies may be benign, others can be hazardous, potentially indicating system faults, security breaches, or other critical issues. The identification of these abnormal patterns is essential as it can provide valuable insights and enable timely interventions in various fields \cite{ramchandran2018unsupervised}.

Anomaly detection plays a critical role across diverse sectors, including healthcare, cybersecurity, fraud detection, IoT, industrial systems, and critical infrastructure. In clinical and biomedical contexts, class imbalance-aware federated imaging \cite{khaniki2025class}, secure cardiac monitoring \cite{ahmadi2025multiheart}, gait abnormality detection in Parkinson's patients \cite{alazeb2024effective}, hospital network EHR analysis \cite{niu2025anomaly}, and wireless body area network security \cite{islam2025securing} highlight the demand for interpretable and privacy-preserving solutions. In industrial and computing domains, AI-based fault detection for power systems \cite{behnam2024decentralized}, real-time anomaly detection \cite{abibulaiev2026context}, TinyML-driven IoT monitoring \cite{katib2025safeguarding}, and evolutionary approaches to industrial anomaly detection \cite{zeng2025evolutionary} underscore the importance of scalable and resource-efficient methods for distributed edge environments.

With the rise of data mining, the demand for robust anomaly detection mechanisms has intensified. Traditional tools like intrusion detection systems (IDS), malware scanners, and network monitoring software typically employ rule-based detection methods, comparing incoming data traffic against predefined patterns or rules to identify potential threats \cite{zhong2022bidirectional}. While effective for known issues, these methods can become outdated and less effective as data patterns evolve and become increasingly complex, necessitating the exploration of more adaptive and intelligent detection strategies.

These challenges are amplified in resource-constrained devices where memory and latency are critical. To address these challenges, modern approaches in anomaly detection increasingly rely on data mining techniques. These techniques involve analyzing large datasets to uncover patterns and associations that may indicate anomalies. Some common methods include clustering, where data is grouped, and classification, where data is divided into categories based on pre-labeled training data. While cluster-based anomaly detection is useful for identifying unfamiliar attacks by segmenting data into smaller subsets, classification methods are often more precise but may struggle with high-volume data due to scalability issues \cite{hu2021real}.

The field of anomaly detection has evolved considerably with the integration of advanced data mining techniques, such as temporal mining, outlier detection, and association rule mining. These methodologies work in synergy to improve the precision and speed of detecting anomalies, particularly in dynamic and high-volume environments. For example, in cybersecurity, real-time network anomaly detection systems are critical for analyzing traffic patterns and identifying suspicious activities like unauthorized access or data breaches \cite{ goswami2024ai}. In healthcare, advanced anomaly detection techniques are utilized to analyze medical imaging data, such as detecting subtle abnormalities in MRI or CT scans that may indicate early-stage diseases \cite{ghajari2024hybrid}. Additionally, video anomaly detection systems are increasingly deployed in security and surveillance, where they are used to identify unusual behaviors in real-time across public transportation hubs, airports, and other sensitive locations \cite{zhong2022cascade,verma2024real,rezaee2024survey}.

Despite the progress in anomaly detection, several challenges remain. The development of versatile algorithms that can adapt to different scenarios and handle large volumes of data in real time is a significant hurdle. Traditional rule-based algorithms often lack the flexibility to adapt to new or changing data patterns, rendering them ineffective in dynamic environments. In contrast, nonparametric learning algorithms offer more adaptability, allowing them to respond to evolving data by learning from previous instances.

Supervised classification learning, a popular approach in machine learning, has gained considerable attention for its effectiveness in anomaly detection. Algorithms such as decision trees, fuzzy logic systems, and support vector machines have been widely used, with artificial neural networks showing particular promise due to their ability to model complex relationships in data \cite{purarjomandlangrudi2014data}. However, challenges such as poor generalization, local convergence issues, and structural difficulties in neural networks still need to be addressed.

To overcome these limitations, hybrid approaches that combine supervised and unsupervised techniques are being explored \cite{shon2007hybrid}. These methods aim to reduce false alarm rates and improve detection accuracy, though they come with their own set of challenges, including complexity in algorithm design, the need for diverse training samples, and difficulties in ensuring scalability. As research in this area continues to evolve, there is significant potential for further advancements in anomaly detection, particularly in improving the scalability, real-time processing, and interpretability of detection models. These developments are crucial for enhancing the effectiveness of anomaly detection across various applications and ensuring the reliability of systems in increasingly complex and data-rich environments. These methods can be resource-intensive, requiring substantial computational power and large amounts of labeled data. Furthermore, they often struggle in environments characterized by high dimensionality, noise, and the need for real-time analysis.

This is where Hyperdimensional Computing (HDC) comes into play. HDC, also known as vector symbolic architecture, is a novel computational paradigm inspired by the way the human brain processes information. Unlike conventional computing methods that rely on precise numerical representations, HDC represents data using high-dimensional binary vectors, also known as hypervectors. These hypervectors can encode complex patterns in a way that is both robust to noise and computationally efficient \cite{kanerva2009hyperdimensional, ge2020classification}. One of the key advantages of HDC is its ability to perform operations on these vectors that mimic cognitive processes, such as associative memory and reasoning, enabling rapid learning and inference.

The importance of HDC in the context of anomaly detection cannot be overstated. Traditional anomaly detection methods often require extensive training and are sensitive to noise and variability in the data. In contrast, HDC's use of high-dimensional spaces allows for the natural separation of normal and anomalous patterns, even in the presence of noise. This makes HDC particularly well-suited for resource-constrained environments, such as IoT devices and embedded systems, where computational efficiency and robustness are paramount \cite{wang2023hyperdetect}. Additionally, HDC's ability to generalize from limited data makes it an attractive option in scenarios where labeled anomalies are scarce.

Applications of anomaly detection using HDC span multiple domains. In cybersecurity, HDC can be employed to detect unusual patterns in network traffic, potentially identifying security breaches in real-time \cite{wang2023hyperdetect,wang2023late,ma2023robust}. In industrial settings, HDC can monitor sensor data from machinery, identifying early signs of equipment failure and thus preventing costly downtime \cite{mitrokhin2019learning,gungor2022res}. In healthcare, HDC can assist in monitoring patient data for signs of irregularities that may indicate the onset of a medical condition \cite{pale2023hyperdimensional,pale2022hyperdimensional}. Despite these advantages, the application of HDC in anomaly detection is not without limitations. One challenge is the need to carefully design the encoding of data into hypervectors, as poor encoding can lead to suboptimal performance. Additionally, while HDC is inherently robust to noise, it may still struggle with very subtle anomalies that are close to the normal data distribution \cite{wilson2023hyperdimensional}.

The relationship between anomaly detection and HDC is symbiotic, with each enhancing the capabilities of the other. Anomaly detection benefits from HDC's ability to efficiently handle high-dimensional data and its robustness to noise, while HDC provides a framework within which complex data patterns can be effectively modeled and analyzed. Recent work has demonstrated HDC's symbolic and interpretable properties in cybersecurity reasoning \cite{zakeri2025enabling}, highlighting the broader applicability of brain-inspired computing paradigms. Our work introduces D2H-AD, which integrates distance-based similarity and density-aware scoring within the HDC encoding framework. By combining these complementary metrics, the proposed method aims to better characterize anomalous patterns in imbalanced and high-dimensional datasets. Empirical evaluation on benchmark datasets demonstrates the effectiveness of this hybrid approach across diverse anomaly detection scenarios.

The following research contributions are reported in this paper.
\begin{enumerate}
\item We introduce D2H-AD, an unsupervised anomaly detection method named after its core components: Density, Distance metrics, and Hyperdimensional Computing, combined for Anomaly Detection. This approach demonstrates competitiveness with supervised methods, all while eliminating the need for extensive training time. The elimination of lengthy training is achieved through Hyperdimensional Computing's (HDC) ability to represent data as high-dimensional binary vectors, allowing for direct computation of density and distance metrics without the iterative optimization processes typically required by traditional machine learning models. By leveraging these pre-encoded hypervectors, D2H-AD efficiently computes anomaly scores without relying on large-scale labeled datasets or time-intensive model training.

\item We propose a novel anomaly detection method that combines density and distance metrics within a Hyperdimensional Computing (HDC) framework. The innovation lies in encoding the dataset into high-dimensional binary vectors, which inherently capture data relationships and patterns. The anomaly detection process is then driven by a two-step approach: first, the encoded hypervectors enable efficient computation of density and distance metrics without the traditional computational overhead; second, an anomaly score is calculated based on these metrics, leveraging the high-dimensional space to naturally separate normal and anomalous patterns. This streamlined yet effective process enhances detection accuracy while suggesting computational efficiency, even in high-dimensional and noisy datasets.

\item We evaluated the performance of the proposed D2H-AD model across five diverse datasets \cite{rayana2016outlier} and compared it with several well-established and recent methods. The results, assessed using F1 score and ROC-AUC metrics, consistently showed that our model outperformed the baseline techniques in all cases. This demonstrates the effectiveness and robustness of our approach in handling various anomaly detection tasks, offering a significant improvement over existing methods. \textcolor{white}{\cite{shul2023noise,sato2024high,li2016anomaly,liu2008isolation,he2020exploring,tang2015outlier,amrouch2023beyond, wang2021brief, wang2022odhd, ghajari2025network,ghajari2025intrusion,xu2025ifodhd,xu2024multi, wang2022poisonhd, wang2022real}}
\end{enumerate}

We structure the subsequent sections of this paper as follows: Section 2 offers a detailed review of related work on Hyperdimensional Computing in anomaly detection. Section 3 details the proposed methodology. In Section 4, we present the results of our empirical evaluation, demonstrating the validity and effectiveness of our approach. Finally, Section 5 offers concluding remarks.

\begin{table*}%[h]
\centering
\caption{Comparison of Existing Anomaly Detection Methods}
\label{tab1:comparison}
\begin{tabular}{|m{4cm}|m{6cm}|m{6.5cm}|}
\hline
\textbf{Method} & \textbf{Strengths} & \textbf{Limitations} \\
\hline
Supervised Deep Learning (e.g., CNN) \cite{shul2023noise} & High accuracy on labeled data; learns complex patterns & Requires large labeled datasets; poor generalization to unseen anomalies; high training cost \\
\hline
Semi-supervised (e.g., GAN)  \cite{sato2024high} & Leverages small labeled set; models non-linear boundaries well & Sensitive to label quality; unstable adversarial training; high resource usage \\
\hline
Unsupervised (e.g., OCSVM, Isolation Forest, Autoencoder)  \cite{li2016anomaly,liu2008isolation,he2020exploring} & No label requirement; suitable for real-world scenarios & Limited performance on complex/high-dimensional data; may overlook subtle anomalies \\
\hline
Statistical Methods (e.g., GMM)  \cite{tang2015outlier} & Interpretable; efficient on structured, low-dimensional data & Assumes specific distribution; not robust for heterogeneous or real-time streams \\
\hline
HDC-based Methods (e.g., HDAD, ODHD, IFODHD, PoisonHD) \cite{wang2021brief,wang2022odhd,amrouch2023beyond,xu2025ifodhd,wang2022poisonhd,christopher2021minority,xu2024multi} & Fast, lightweight, interpretable; robust to noise and imbalance; explainable; efficient on small data; hardware/edge friendly & Encoding strategies require task-specific design choices (e.g., quantization levels, projection dimensions, feature selection thresholds vary across domains); limited public benchmarks; evolving defense mechanisms \\
\hline
Proposed Approach (D2H-AD) & Combines dual-domain HDC encoding with optimized feature selection and PoisonHD-style defense; explainable and scalable for edge/IoT & Under evaluation; needs further benchmarking and generalization across diverse datasets \\
\hline
\end{tabular}
\end{table*}
%$$$$$$$$$$$$$$$$$$$$$$$$$$$$$$$$$$$$$$$$$$$$$$$$$$$$$$
\begin{table*}%[h]
\centering
\caption{Practical comparison of anomaly detection approaches across deployment-related aspects.}
\label{tab2:practical_comparison}
\begin{tabular}{|p{4cm}|c|c|c|c|}
\hline
\textbf{Method} & \textbf{Label Requirement} & \textbf{Handles HD Data} & \textbf{Computational Complexity} & \textbf{Edge/IoT Friendly} \\
\hline
Supervised DL (e.g., CNN) & High & Yes & High & No \\
\hline
Semi-supervised (e.g., GAN) & Medium & Yes & High & No \\
\hline
OCSVM / Isolation Forest & None & Partially & Low--Medium & Moderate \\
\hline
Autoencoder & None & Yes & Medium--High & Moderate \\
\hline
Statistical (e.g., GMM) & None & No & Low & Yes \\
\hline
HDC-based {\fontsize{6pt}{7pt}\selectfont(HDAD, ODHD, IFODHD)} & None & Yes & Low & Yes \\
\hline
HDC + Defense (e.g., PoisonHD) & None & Yes & Low--Medium & \makecell{Yes\\resilient with memory sanitization} \\
\hline
\textbf{Proposed (D2H-AD)} & None & Yes & \textbf{Very Low} & \textbf{Suitable (expected real-time)} \\

\hline
\end{tabular}

\smallskip
\noindent\begin{minipage}{\linewidth}\footnotesize
\textbf{Notes.} 
\emph{Label Requirement: None} = method runs unsupervised without labeled 
anomalies.  
\emph{Handles HD Data: Yes} = binary hypervectors preserve robustness in 
high dimensions.  
\emph{Computational Complexity} ratings are based on algorithmic operation 
counts (XOR, addition) and memory requirements (linear in $m$, $n$, $D$).
\emph{Edge/IoT Friendly} and \emph{Real-time suitability} refer to algorithmic 
properties (binary operations, linear memory) that suggest hardware efficiency; 
actual hardware implementation and benchmarking are not included in this work.
\end{minipage}
\end{table*}

\section{Related Work}

Anomaly detection is a cornerstone in modern artificial intelligence and data mining, playing a pivotal role in diverse domains such as cybersecurity, healthcare diagnostics, finance, industrial monitoring, and smart grid systems. The primary objective is to identify unusual patterns or behaviors that differ significantly from the norm, which often correspond to critical faults, fraudulent activities, or emergent system threats. Depending on the availability of labeled data and the complexity of the underlying problem, anomaly detection techniques are broadly classified into supervised, semi-supervised, and unsupervised categories.

Supervised approaches rely on a dataset that includes labeled examples of both normal and anomalous behavior. These models, especially deep learning architectures like Convolutional Neural Networks (CNNs), have been extensively applied in tasks such as network intrusion detection, medical image analysis, and fraud detection \cite{shul2023noise}. They can achieve high performance when trained on large, high-quality datasets. Nevertheless, acquiring such labeled datasets is non-trivial, particularly in domains where anomalies are rare or costly to label. Additionally, supervised models may exhibit poor generalization to previously unseen anomaly types, a limitation often referred to as the open-set recognition problem.

Semi-supervised anomaly detection methods typically assume that the available data is predominantly normal, with very limited labeled anomalies. These models learn to characterize the normal data distribution and identify deviations as potential outliers. Generative Adversarial Networks (GANs) have gained popularity in this space for their ability to generate synthetic samples and refine decision boundaries \cite{sato2024high}. While GAN-based approaches offer flexibility and often outperform traditional methods, they are prone to training instability and can be sensitive to class imbalance or noisy labels.

Unsupervised methods are the most flexible in real-world settings, as they do not require labeled data. Techniques like One-Class Support Vector Machines (OCSVM), Isolation Forests, and Autoencoders have become widely adopted \cite{li2016anomaly,liu2008isolation,he2020exploring}. OCSVM creates a decision boundary around normal data, Isolation Forests randomly partition data and identify anomalies as those requiring fewer splits, and Autoencoders reconstruct input data to measure reconstruction errors as indicators of anomalous behavior. While these approaches are scalable and generally effective, they often struggle with high-dimensional data, temporal dependencies, and subtle anomalies embedded within complex patterns. These limitations have also been documented in hardware security contexts, where OC-SVM and reconstruction-based Autoencoders proved fragile under process variation and adversarial perturbations \cite{ghimire2026adversarial}. Recent advances in neuromorphic computing have also explored evolving spiking neural networks for unsupervised anomaly detection in streaming data, demonstrating competitive performance with reduced computational overhead \cite{rehan2025hybrid, rehan2025hyperparameters}.

While deep Autoencoders are widely used for anomaly detection due to their ability to capture nonlinear structures, they typically require large parameterized networks, iterative gradient descent, and significant memory resources \cite{vincent2008extracting}. In contrast, HDC-based methods operate with lightweight binary operations, achieving comparable accuracy with far lower computational overhead \cite{chang2023recent}. Moreover, unlike Autoencoders that behave as opaque models, HDC offers feature-level interpretability through prototype hypervectors, making it more suitable for safety-critical and resource-constrained applications.

Autoencoders detect anomalies through reconstruction error \cite{torabi2023practical}, which can fail when anomalies lie on or near the learned data manifold, whereas HDC methods like D2H-AD explicitly model local density and relative distances in high-dimensional space, enabling detection of subtle anomalies that exhibit normal reconstruction patterns but deviate in their neighborhood structure. The single-pass encoding of HDC eliminates the need for iterative training epochs, while the distributed representation across high-dimensional binary vectors provides inherent robustness to noise and missing features that would require explicit regularization in autoencoder architectures. These theoretical 
advantages are reflected in our comparative analysis (Tables~\ref{tab1:comparison} 
and~\ref{tab2:practical_comparison}), where HDC-based methods demonstrate superior 
edge-deployment suitability and lower computational complexity than autoencoders.

In addition to learning-based techniques, statistical models such as the Gaussian Mixture Model (GMM) with Expectation Maximization have also been explored for anomaly detection tasks \cite{tang2015outlier}. While GMMs are mathematically elegant and computationally efficient, their reliance on parametric assumptions, such as the data following a Gaussian distribution, limits their effectiveness in complex, high-variance, or multimodal datasets.

Recent advancements have seen the emergence of Hyperdimensional Computing (HDC), inspired by principles of brain functionality and cognitive computing. HDC operates on high-dimensional vectors known as hypervectors, which represent data in a distributed and noise-resilient manner \cite{amrouch2023beyond}. Unlike conventional numerical computation, HDC enables fast, parallelizable, and interpretable operations like binding, bundling, and permutation. Due to its low computational complexity, inherent robustness, and ability to generalize from few examples, HDC has become an attractive framework for anomaly detection, particularly in resource-constrained environments like embedded systems and edge devices.

\subsection{Hyperdimensional Computing Fundamentals}

\subsubsection{Core Principles and Vector Generation}
Hyperdimensional Computing (HDC) represents information using high-dimensional binary or bipolar vectors (hypervectors) of dimension $D$, typically $D \geq 10{,}000$. The computational model relies on three fundamental operations that preserve semantic relationships in the hyperdimensional space:
\begin{itemize}
    \item \textbf{Binding ($\otimes$):} Combines two hypervectors to create a dissimilar composite. For binary vectors, binding is implemented via element-wise XOR: $\mathbf{h} = \mathbf{h}_i \otimes \mathbf{h}_j$ where $\mathbf{h}[k] = \mathbf{h}_i[k] \oplus \mathbf{h}_j[k]$. For bipolar vectors $\{-1,+1\}$, element-wise multiplication is used. Binding is approximately invertible: $\mathbf{h}_i \otimes (\mathbf{h}_i \otimes \mathbf{h}_j) \approx \mathbf{h}_j$.
    \item \textbf{Bundling ($+$):} Aggregates multiple hypervectors into a composite representation that is similar to all constituents. For binary vectors: $\mathbf{H} = \text{MAJ}(\mathbf{h}_1 + \mathbf{h}_2 + \ldots + \mathbf{h}_n)$ where MAJ binarizes by majority vote. For bipolar vectors, element-wise addition followed by sign function is used.
    \item \textbf{Permutation ($\rho$):} Rearranges vector elements to encode sequential or positional information, typically via circular shift: $\rho(\mathbf{h}) = [\mathbf{h}[D], \mathbf{h}[1], \mathbf{h}[2], \ldots, \mathbf{h}[D-1]]$.
\end{itemize}

\subsubsection{Encoding Strategies}
HDC encoding transforms raw features into hypervectors while preserving semantic relationships. Common strategies include:
\begin{itemize}
    \item \textbf{Level Hypervectors:} Continuous values are quantized into $K$ levels, each assigned a hypervector. Adjacent levels have similar hypervectors (Hamming distance $\approx D/K$), creating a smooth similarity gradient \cite{kanerva2009hyperdimensional}.
    \item \textbf{Random Projection:} Dense vectors are projected into high-dimensional space using random matrices, leveraging the Johnson-Lindenstrauss lemma for dimensionality expansion while preserving distances.
    \item \textbf{Sparse Distributed Representation:} Inspired by fruit fly olfactory circuits, sparse hypervectors ($\approx 1$--$5\%$ active bits) can enhance separability and reduce computational cost in specific applications.
    \item \textbf{N-gram Encoding:} For sequential data, overlapping n-grams are encoded by binding character/token hypervectors with position-dependent permutations.
\end{itemize}

\subsubsection{Orthogonality and Randomness}
The effectiveness of HDC critically depends on the quasi-orthogonality of randomly generated base hypervectors. For $D = 10{,}000$ and Bernoulli$(0.5)$ generation, the Hamming distance between independent hypervectors concentrates around $D/2 \pm \sqrt{D}$ with exponentially high probability. This statistical orthogonality ensures that binding operations produce dissimilar outputs and bundling preserves distinguishability. Recent work has explored:
\begin{itemize}
    \item Deterministic orthogonal codes (e.g., Hadamard matrices, Discrete Fourier Transform bases)
    \item Hardware-efficient generation using Linear Feedback Shift Registers (LFSRs)
    \item Biologically-plausible sparse random projections
    \item Structured random matrices for reduced generation complexity
\end{itemize}

\subsubsection{Platform Efficiency and Hardware Implementations}
HDC's appeal for edge computing stems from its intrinsic hardware 
efficiency:
\begin{itemize}
    \item \textbf{Bit-serial Processing:} Binary operations (XOR, 
    popcount) map directly to digital logic without floating-point 
    units
    
    \item \textbf{In-Memory Computing:} Associative memory architectures 
    (e.g., resistive RAM, memristors) can perform HDC operations with 
    reduced data movement
    
    \item \textbf{FPGA Implementations:} Parallel bit-wise operations 
    achieve sub-microsecond inference latency
    
    \item \textbf{Neuromorphic Hardware:} Spiking implementations on 
    Intel Loihi and IBM TrueNorth demonstrate energy efficiency gains 
    of $10$--$100\times$ over GPU baselines \cite{wu2022brain}
\end{itemize}
Recent benchmarking studies show HDC achieving $1000\times$ speedup 
and $100\times$ energy reduction compared to deep neural networks on 
embedded platforms, with accuracy within $1$--$3\%$ on classification 
tasks.

\subsubsection{Recent Surveys and Theoretical Advances}
Recent comprehensive surveys have characterized HDC's theoretical 
foundations and practical deployment 
\cite{chang2023recent,kleyko2022vector,stock2024hyperdimensional}:
\begin{itemize}
    \item Capacity analysis showing that $D$-dimensional binary 
    hypervectors can reliably store $O(D^2/\log D)$ patterns
    
    \item Robustness guarantees demonstrating graceful degradation 
    under up to $D/4$ bit errors
    
    \item Convergence proofs for iterative HDC learning showing 
    $O(\log n)$ training iterations for $n$-class problems
    
    \item Cross-domain benchmarking across language, vision, sensor 
    fusion, robotics, and bioinformatics applications
\end{itemize}

%$$$$$$$$$$$$$$$$$$$$$$$$$$$$$$$$$$$$$$$$$$$$$$$$$$$$$$
\begin{table*}%[h]
\centering
\caption{Applications of HDC-based Anomaly Detection Across Domains}
\label{tab3:hdc_applications}
\begin{tabular}{|m{3.3cm}|m{4cm}|m{4cm}|m{4.5cm}|}
\hline
\textbf{Domain} & \textbf{Application} & \textbf{HDC Method / Reference} & \textbf{Key Advantages} \\
\hline
\textbf{Cybersecurity (IoT)} & Intrusion detection using NSL-KDD dataset & HDC-based IDS \cite{ghajari2025network,ghajari2025intrusion} & High accuracy (up to 99.54\%); efficient with high-dimensional features; robust to diverse attack types \\
\hline
\textbf{Data Imbalance at Edge} & Threat detection with minority class oversampling & HDC + EG-SMOTE \cite{christopher2021minority} & Overcomes class imbalance; outperforms standard resampling methods \\
\hline
\textbf{Healthcare / Biomedicine} & Abnormal heartbeat classification from ECG signals & HDC + AUC-based feature selection \cite{xu2024multi} & High accuracy in multi-class settings; interpretable for clinical use; resilient to noise \\
\hline
\textbf{Smart Grid / Energy} & Real-time anomaly detection in power meter data & Real-time HDC \cite{wang2022real} & No preprocessing needed; low-latency; edge-device deployable \\
\hline
\textbf{Embedded AI / TinyML} & General anomaly detection for edge and low-power devices & ODHD, IFODHD \cite{wang2022odhd,xu2025ifodhd} & Low memory footprint; scalable encoding; suitable for TinyML and mobile deployment \\
\hline
\end{tabular}
\end{table*}
%$$$$$$$$$$$$$$$$$$$$$$$$$$$$$$$$$$$$$$$$$$$$$$$$$$$$$$
\subsection{HDC-Based Anomaly Detection Methods}
\label{subsec:hdc_anomaly_methods}

Having established the theoretical foundations of Hyperdimensional 
Computing, we now examine how these principles have been applied to 
anomaly detection across diverse application domains.

The HDAD and ODHD frameworks demonstrate how HDC can be applied to anomaly detection tasks with minimal reliance on labeled data \cite{wang2021brief,wang2022odhd}. ODHD, a one-class model, removes the need for decoder architectures and leverages HDC's strength in handling high-dimensional spaces efficiently. In practice, ODHD outperforms traditional models on classification accuracy and runtime while maintaining low memory and energy consumption.

Recent research has explored the use of Hyperdimensional Computing for network-based anomaly detection, particularly in IoT environments. For instance, in \cite{ghajari2025network}, HDC was applied to the NSL-KDD dataset for intrusion detection, demonstrating superior performance with an accuracy of 91.55\%. Similarly, \cite{ghajari2025intrusion} presented an HDC-based framework that accurately classified various attack types in IoT networks, achieving 99.54\% accuracy. These studies highlight the efficiency and robustness of HDC in handling high-dimensional cybersecurity tasks, and motivate the extension of HDC-based methods beyond network intrusion to broader anomaly detection contexts, as pursued in our proposed D2H-AD.

To address the issue of data imbalance at the IoT edge, \cite{christopher2021minority} proposed a hybrid approach combining HDC with Enhanced Geometric SMOTE (EG-SMOTE). This method achieved improved accuracy in threat detection and significantly outperformed conventional oversampling and classification techniques on benchmark cybersecurity datasets, demonstrating the synergy of HDC with data-level balancing techniques.

Further improvements have been realized through the integration of feature selection. IFODHD \cite{xu2025ifodhd} combines the Minimum Redundancy Maximum Relevance (mRMR) feature selection method with HDC's encoding process. This reduces feature dimensionality and redundancy, improving model performance across diverse datasets. The approach not only enhances classification metrics (accuracy, F1 score, ROC-AUC) but also significantly decreases model latency and memory footprint, making it ideal for applications in wearable devices, autonomous systems, and mobile platforms.

In the healthcare domain, \cite{xu2024multi} applied HDC to classify abnormal cardiac rhythms in arrhythmia datasets. By incorporating an AUC-based feature selection process into the HDC pipeline, the proposed system outperformed classical models like SVM and MLP, confirming HDC's potential in multi-class and imbalanced clinical data scenarios.

In parallel, HDC proved effective in real-time industrial applications. In \cite{wang2022real}, HDC was applied to electrical load anomaly detection in smart grids. The model operated directly on raw meter data without preprocessing, offering superior performance over traditional ML/DL approaches in terms of speed, scalability, and detection accuracy, particularly in imbalanced data environments. Its successful deployment on a mid-range laptop underscored its efficiency and readiness for integration into edge computing systems such as smart meters, energy-efficient controllers, and IoT gateways.

Moreover, HDC models are uniquely positioned for explainable AI (XAI). Their algebraic nature allows intuitive tracing of decisions through hypervector operations. This property is particularly useful in critical systems such as healthcare or autonomous vehicles, where interpretability and trust are paramount. Recent studies have proposed integrating causal analysis with HDC, enhancing the transparency of detected anomalies and offering insights into their underlying causes.

Security concerns have also emerged with the adoption of HDC in sensitive systems. PoisonHD \cite{wang2022poisonhd} presents the first targeted attack model for HDC systems, using a label-flipping strategy guided by confidence ranking. The study reveals that despite HDC's robustness, it can still be vulnerable to poisoning attacks. To mitigate this, PoisonHD introduces an HDC-specific defense mechanism based on data sanitization using verified memory modules. This sanitization filters poisoned samples before training and has been shown to restore performance to near-baseline levels. The findings emphasize the growing importance of adversarial resilience in lightweight, deployable AI.

While HDC-based methods offer computational efficiency, their encoding strategies often require task-specific design choices. For instance, HDAD \cite{wang2021brief} relies on explicit selection of quantization levels and base hypervector initialization schemes that differ between continuous sensor data (e.g., automotive CAN bus signals) and discrete categorical features. Similarly, ODHD \cite{wang2022odhd} requires dataset-dependent tuning of the projection dimension and the number of training epochs for prototype learning, which varies substantially between image-based tasks (e.g., MNIST) and tabular datasets. IFODHD \cite{xu2025ifodhd} further introduces feature selection hyperparameters (e.g., mRMR thresholds) that must be calibrated per domain to balance redundancy reduction and information retention. In contrast, the proposed D2H-AD adopts a unified encoding scheme based on fixed base and level hypervectors that is applied in a consistent manner across datasets, while requiring only a small set of generally applicable hyperparameters (e.g., $D$, $K$, and the $dc$ percentile), whose impact is analyzed through sensitivity experiments (Section~\ref{Hyperparameter}, Table~\ref{tab10E}).

The evolution of HDC-based models has also inspired hybrid techniques that combine symbolic reasoning, graph representations, and neural modules within hyperdimensional frameworks. Such architectures are promising for tasks that require both structured knowledge and data-driven adaptation, such as multimodal anomaly detection and spatio-temporal modeling. Integrating these hybrid mechanisms into anomaly detection pipelines can further improve performance, adaptability, and generalization across heterogeneous datasets.

Beyond these application-specific HDC methods, recent systematic surveys and benchmarking studies have established HDC's theoretical foundations and practical trade-offs across diverse domains. Cross-domain evaluations demonstrate that HDC achieves competitive accuracy within 1--5\% of deep learning baselines while offering 10--100$\times$ improvements in energy efficiency and inference latency on embedded platforms. These comprehensive studies emphasize the critical importance of encoding design choices, with level-based quantization schemes, as employed in D2H-AD, demonstrating superior performance on continuous-valued tabular data compared to random projection or sparse encoding alternatives. This body of empirical evidence motivates our design decisions and provides context for the unified encoding strategy adopted in the proposed methodology.

Tables~\ref{tab1:comparison} and~\ref{tab2:practical_comparison} provide a comprehensive comparison of existing anomaly detection methods from two complementary perspectives. Table~\ref{tab1:comparison} summarizes the key strengths and limitations of each approach, highlighting the unique contributions of HDC-based models and our proposed method. In contrast, Table~\ref{tab2:practical_comparison} focuses on practical deployment aspects such as label dependency, suitability for high-dimensional data, computational cost, and applicability in resource-constrained environments like IoT and edge computing.

Furthermore, Table~\ref{tab3:hdc_applications} illustrates the broad applicability of HDC-based anomaly detection across diverse domains such as cybersecurity, healthcare, energy systems, and embedded AI. This table showcases the flexibility, interpretability, and performance advantages of HDC-based methods. Together, these comparisons clearly demonstrate the novelty, efficiency, and real-world relevance of our proposed approach.

In summary, the landscape of anomaly detection is undergoing a significant transformation. Traditional methods, while still valuable, are increasingly being augmented or replaced by lightweight, interpretable, and scalable alternatives like HDC. From efficient encoding and real-time inference to adversarial defense and cross-domain applicability, HDC models such as ODHD, IFODHD, and PoisonHD have laid a solid foundation. Building upon this trajectory, our proposed method, D2H-AD, introduces a dual-domain HDC-based architecture that integrates optimized encoding, feature selection, and embedded defense mechanisms. It aims to deliver robust, adaptive, and secure anomaly detection suitable for next-generation edge AI and real-time monitoring systems.
\begin{figure*}%%[!t]
    \centering
    \includegraphics[width=\textwidth]{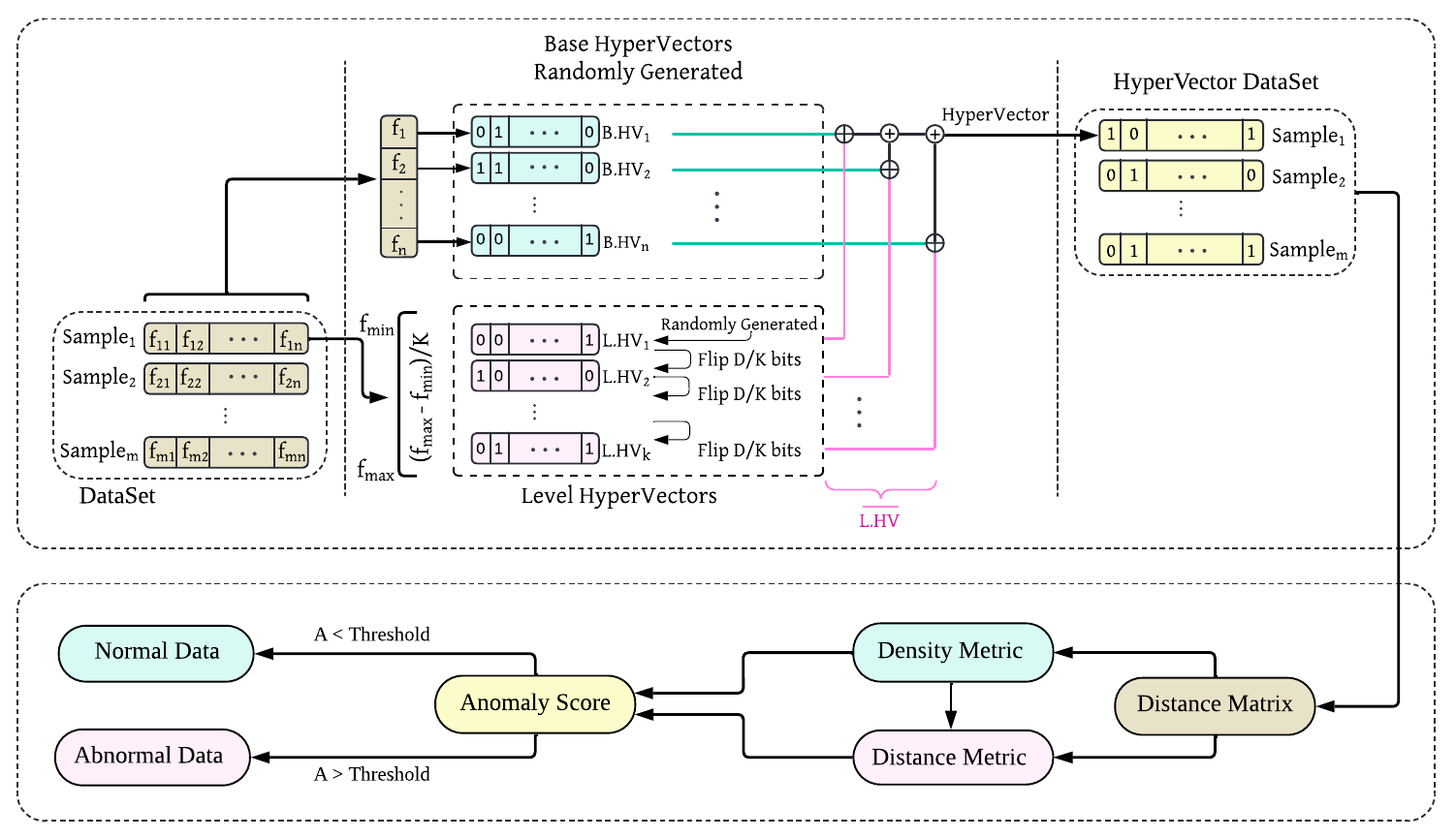}
    \caption{\textbf{D2H-AD Framework.} 
\textit{Top:} The encoding phase transforms raw input features into binary hypervectors through base and level hypervector binding and bundling operations. 
\textit{Bottom:} The anomaly detection phase computes density and distance metrics in the hyperdimensional space, where the hybrid anomaly score $A=\delta/\rho$ amplifies isolated low-density outliers.}

    \label{fig1}
\end{figure*}
\section{Methodology}
In this section, we present the D2H-AD methodology, a novel approach for anomaly detection that leverages the capabilities of Hyperdimensional Computing (HDC). The overall framework of the proposed method is depicted in Figure \ref{fig1}, which is structured into two distinct phases: the encoding phase and the anomaly detection phase.

The upper part of Figure \ref{fig1} illustrates the encoding phase, where raw input data is transformed into high-dimensional vector representations. This phase plays a critical role in capturing the underlying characteristics of the data through distributed and dense representations in the hyperdimensional space, ensuring robustness and scalability.

The lower part of the figure represents the anomaly detection phase, where the encoded high-dimensional vectors are analyzed using our hybrid strategy. This phase combines distance-based and density-based measures to accurately identify anomalies by leveraging the unique properties of the high-dimensional space.

In the subsequent subsections, we provide a detailed explanation of each phase, elaborating on the methodologies, algorithms, and mechanisms employed in both the encoding and anomaly detection stages. The step-by-step description highlights how D2H-AD efficiently integrates HDC principles with advanced anomaly detection strategies to enhance accuracy and performance.

\subsection{Encoding}
The first step in encoding the dataset into hypervectors is to ensure that all the information from the original data points, including both feature values and feature indices, is retained in the high-dimensional space. Each data point consists of $n$ features, and the goal is to map these features into a binary hypervector of dimension $D = 10,000$.
\subsubsection*{Step 1: Base Hypervector Assignment}
For each feature $f_j \in \{f_1, \dots, f_n\}$ in the dataset, we generate a 
unique base hypervector $B.HV_j \in \{0,1\}^D$. Each dimension is independently 
sampled from a Bernoulli distribution with probability $p = 0.5$, i.e., 
$B.HV_j[d] \sim \text{Bernoulli}(0.5)$ for $d \in \{1, \dots, D\}$. This ensures 
that base hypervectors are approximately orthogonal with expected Hamming distance
\[
\mathbb{E}[d_H(B.HV_i, B.HV_j)] = \frac{D}{2} \quad \text{for } i \neq j,
\]
with high-probability concentration guaranteed by Hoeffding's inequality when 
$D \ge 10{,}000$. In our implementation, we use Python's 
\texttt{numpy.random.default\_rng()} with a fixed seed for reproducibility, 
which implements the PCG64 pseudorandom number generator.
%%%%%%%%%%%%%%%%%%%%%%%%%%%%%%%%%%%%%%%%%%%%%%%%%%%%%%%%%%%%%%%%%%%%%%%%
%&&&&&&&&&&&&&&&&&&&&&&&&&&&&&&&&&&&&&&&&&&&&&&&&&&&&&&&&&&&&&&&&&&&&&&&&

\begin{table}%%[htbp]
\centering %\begin{center}
\begin{tabular}{l}\hline
\textbf{Algorithm 1:} Pseudocode for D2H-AD - Encoding ~~~~~~~~~~~~~~~~~~~~~~~\vspace{0.1cm}
 \\ \hline
%\hspace{0.2cm} 
\textbf{Input:} Dataset with $m$ samples and $n$ features\\ 
\textbf{Output:} HyperVector Dataset with $m$ samples\\
$D$: length of Hyper vector\\   
$K$: number of levels \\   
$f_{\text{min}}$ and $f_{\text{max}}$: the smallest and largest features among all data points. \\   
%$m$: number of samples \\   
\textbf{Base HyperVector:}\\
for each feature $j$ in $n$ features: \\  \hspace{0.2cm}
Create a random Binary Basic HyperVector ($B.HV_j$) of length $D$ \\
\textbf{Level HyperVector:}\\
Split the input feature space \((f_{\text{min}}, f_{\text{max}})\) into $K$ uniform intervals.\\
Create a random Binary Level HyperVector ($L.HV_1$) of length $D$ \\
for $i$ in range $(2:K)$:\\  \hspace{0.2cm}
Create $L.HV_i$ by flipping $D/K$ bits of $L.HV_{i-1}$ \\
Assign each $L.HV$ to its corresponding interval.\\
\textbf{HyperVector Dataset:}\\
for each sample $x$ in $m$ samples:\\  \hspace{0.2cm}
$HV_x$: a HyperVector for sample $x$  \\  \hspace{0.2cm}
for each feature $j$ in the sample:\\  \hspace{0.4cm}
Assign each feature value to its corresponding interval as $\overline{L.HV_j}$ \\ \hspace{0.4cm}
$HV_x$ = $HV_x + B.HV_j \bigoplus \overline{L.HV_j} $ \\
\textbf{Normalize HyperVector Dataset:}\\  %\hspace{0.2cm}
for each sample $x$ in $m$ samples:\\  \hspace{0.2cm}
for $i$ in range ($D$) : \\ \hspace{0.4cm}
if $HV_x[i] <= n/2$ : \\ \hspace{0.6cm}
$HV_x[i]=0$ \\\hspace{0.4cm}
else: \\\hspace{0.6cm}
$HV_x[i]=1$
\\ \hline
\end{tabular}
\label{alg1}
%\end{center}
\end{table}

\begin{table}%%[htbp]
%\caption{Performance Evaluation Considering Different Architectures}
\centering %\begin{center}

\begin{tabular}{l}\hline
\textbf{Algorithm 2:} Pseudocode for D2H-AD - Anomaly Detection ~~~~~~~~~~~\vspace{0.1cm}
 \\ \hline
 \textbf{Input:} HyperVector Dataset (HVD) with $m$ samples \\  
\textbf{Output:} Normal and Abnormal Data Points \\   
\textbf{Distance Matrix:}\\
$DM=1-(HVD*HVD^T)$ \\
\textbf{Density Metric:}\\
$dc$: the 10th percentile threshold of the distance distribution \\
$DenM$: a zero list with length $m$ \\
 $j$ $!=i$ \\
for $i$ in range $(m)$: \\ \hspace{0.2cm}
%count all points closer than $dc$ to the point $i$ \\
for $j$ in range $(m)$:\\  \hspace{0.4cm}
if $DM_{i,j} < dc$ : \\ \hspace{0.6cm}
$DenM_i ++$ \\
\textbf{Distance Metric:}\\
$DisM$: a zero list with length $m$ \\
for $i$ in range $(m)$: \\ \hspace{0.2cm}
Find the least dense point \(j\) that is denser than point \(i\) in $DenM$. \\ \hspace{0.2cm}
$DisM_i = DM_{ij}$\\ \hspace{0.2cm}
if the point $i$ is the densest point:\\ \hspace{0.2cm}
$DisM_i = Max(DM)$ \\
\textbf{Anomaly Score:}\\
$An=DisM/DenM$ \\
\textbf{Detection:}\\
%Threshold \\
%abnormal: an empty list \\
%normal: an empty list \\
for each sample $i$ in $HDV$:\\ \hspace{0.2cm}
if $An_i$ $>$ Threshold: \\ \hspace{0.4cm}
sample $i$ is abnormal \\ \hspace{0.2cm}
else: \\ \hspace{0.4cm}
sample $i$ is normal \\ 
\hline
\end{tabular}
\label{alg2}
%\end{center}
\end{table}
%&&&&&&&&&&&&&&&&&&&&&&&&&&&&&&&&&&&&&&&&&&&&&&&&&&&&&&&&&&&&&&&&&&&&&&
%%%%%%%%%%%%%%%%%%%%%%%%%%%%%%%%%%%%%%

\subsubsection*{Step 2: Level Hypervector Creation}
Next, the range of each feature in the dataset, from the minimum to the maximum 
value, is divided into $K$ quantized intervals. For each of these intervals, a 
corresponding level hypervector is generated. The creation of level hypervectors, 
referred to as $L.HV$, follows a progressive flipping scheme:
\begin{itemize}
    \item The first-level hypervector $L.HV_1$ is generated randomly, with each 
    dimension $L.HV_1[d] \sim \text{Bernoulli}(0.5)$ sampled independently. This 
    initialization ensures statistical independence from base hypervectors while 
    maintaining an expected density of approximately $D/2$ ones.
    
    \item For each subsequent level $\ell \in \{2, \dots, k\}$, we uniformly 
    select $D/k$ random bit positions and flip them in $L.HV_{\ell-1}$ to produce 
    $L.HV_\ell$. This progressive modification continues until we reach the 
    $k^{\text{th}}$ level hypervector $L.HV_k$.
\end{itemize}
As a result, hypervectors corresponding to adjacent intervals exhibit high 
similarity (small Hamming distance), while level hypervectors for distant 
intervals (e.g., $L.HV_1$ and $L.HV_k$) have an expected Hamming distance of 
approximately $D(k-1)/k$, reflecting their semantic dissimilarity. This design 
ensures that data points with similar feature values are encoded with similar 
level hypervectors, naturally preserving the ordinal structure of the original 
data in the hyperdimensional space.
\subsubsection* {Step 3: Mapping Features to Hypervectors}

For each feature in a data point, we map it to the corresponding interval based on its value, and assign the associated level hypervector. To encode both the feature value and its index, we perform an element-wise XOR operation between the feature's $B.HV$ hypervector and its $L.HV$. For example, if a feature  $f_i$  belongs to the  $j^{th}$ interval, its corresponding hypervector is calculated as:
\begin{center}
    $h_i = B.HV_i \oplus L.HV_j$
\end{center} 

\subsubsection*{Step 4: Combining Hypervectors for a Data Point}

After calculating the hypervector for each feature, we sum these feature hypervectors element-wise across all features in the data point to generate the final hypervector for the entire data point. The resulting hypervector may have integer values due to the summation of binary vectors:

\[H = \sum_{i=1}^{n} (B.HV_i \oplus L.HV_i) =\sum_{i=1}^{n} h_i \]

\subsubsection*{Step 5: Binarization Using Majority Function}

To convert the hypervector into a binary format, we apply a majority function. For each dimension of the hypervector, we compare the value against a threshold (typically $n/2$, where  $n$ is the number of features), and binarize the hypervector based on this comparison:

\[H{\prime} = \text{MAJ}(H, n/2)\]

\noindent Where the majority function MAJ is defined as:

\[
H{\prime}_i =
\begin{cases}
1 & \text{if } H_i \geq n/2 \\
0 & \text{otherwise}
\end{cases}
\]

\noindent The final output is a binary hypervector $H{\prime} \in \{0, 1\}^D$, which encodes the entire data point in the high-dimensional space. The details of the encoding procedure used in the proposed method are outlined in Algorithm 1.

\subsubsection*{Randomness Quality and Orthogonality Properties}
While our current implementation relies on standard pseudorandom generation 
(PCG64) without explicit orthogonality verification, the high dimensionality 
($D = 10{,}000$) provides strong probabilistic guarantees. By Hoeffding's 
inequality, the Hamming distance between any two independently generated 
$D$-bit hypervectors concentrates tightly around $D/2$ with deviation bounded 
by $O(\sqrt{D \log(1/\delta)})$ with probability at least $1-\delta$. For 
$D = 10{,}000$, this yields near-orthogonality (approximately 50\% overlap 
$\pm$ 1\%) with probability exceeding $1 - e^{-\Omega(D)}$.

We acknowledge that we have not explored alternative randomness sources or 
performed explicit orthogonality checks in this work. The random vector 
generation is indeed one of the most computationally expensive operations 
in HDC (along with binding and bundling), particularly when generating large 
codebooks. Future work could investigate several promising directions:
\begin{itemize}
    \item Hardware true random number generators (TRNGs) for enhanced 
    unpredictability and security
    \item Structured random matrices (e.g., Fast Hadamard Transform, circulant 
    matrices) to reduce generation cost from $O(nD)$ to $O(n \log D)$
    \item Deterministic quasi-orthogonal codes (e.g., Gold codes, Kasami 
    sequences) for guaranteed minimum distance properties
    \item Biologically-inspired sparse randomness, as observed in fruit fly 
    olfactory encoding, for improved efficiency and interpretability
    \item Explicit orthogonality verification using approximate nearest neighbor 
    algorithms or random projection-based methods
\end{itemize}
These alternatives may reduce computational overhead while maintaining or 
improving separability and robustness properties essential for effective 
hyperdimensional encoding.
%%%%%%%%%%%%%%%%%%%%%%%%%%%%%%%%%%%%%%%%%%%%%%%%%%55

%%%%%%%%%%%%%%%%%%%%%%%%%%%%%%%%%%%%%%%%%%%%%%%%%%%%%%%%%%%%%%%%%%%%%%%%

\subsection{Anomaly Detection}
After encoding the data points into hypervectors, the D2H-AD model employs a hybrid approach to detect anomalies. This process consists of several key components: distance matrix computation, local density estimation, minimum distance calculation to higher density points, anomaly score computation, and final anomaly detection which are explained in detail as follows:

\subsubsection*{Step 1: Distance Calculation}

To begin, we compute a distance matrix between all pairs of encoded hypervectors. For any two hypervectors $x_i$ and $x_j$, we calculate their distance using either the Hamming distance or approximate cosine distance:
\begin{itemize}
    \item Hamming distance:
\end{itemize}
    \[D(x_i, x_j)=\sum(x_i\oplus x_j)\]

where $\oplus$ denotes the XOR operation.
\begin{itemize}
    \item Approximate cosine distance:
   \end{itemize}
\[ D(x_i, x_j)=1-(x_i \cdot x_j) \]
   
where $\cdot$ represents the dot product operation.
These distance measures ensure that we capture both exact and approximate dissimilarities between hypervectors.

\subsubsection*{Step 2: Local Density Estimation}
For each data point $i$, we estimate its local density $\rho_i$ using:
\[ \rho_i = \sum_j \chi(D(x_i, x_j) - dc) \]
where:
\begin{itemize}
    \item $dc$ is a cutoff distance, determined as the $10^{th}$ percentile of all pairwise distances in the dataset.
    \item $\chi(x)$ is an indicator function:
\end{itemize}
\[
\chi(x) =
\begin{cases} 
1, & \text{if } x < 0 \\
0, & \text{otherwise}
\end{cases}
\]
This calculation effectively counts the number of points within the cutoff distance $dc$ from point $i$, indicating the local density around it.

\subsubsection*{Step 3: Minimum Distance Calculation}
For each data point $i$, we compute $\delta_i$, which represents the minimum distance to any other point $j$ that has a higher density:
\[
\delta_i = \min_{j \, : \, \rho_j > \rho_i} D(x_i, x_j)
\]

%\begin{center}
%$\delta_i=\min(D(x_i, x_j))$ for all $j$ where $\rho_j>\rho_i$ 
%\end{center}
For the point with the highest density, $\delta_i$ is defined as the maximum distance to any other point:
\[
\delta_i = \max_{j} D(x_i, x_j)
\]

%\begin{center}
%$\delta_i=\max(D(x_i, x_j))$ for all $j$
%\end{center}
This step ensures that each point is evaluated based on its relative positioning to denser regions in the data space.

\subsubsection*{Step 4: Anomaly Score Computation}
The anomaly score $A_i$ for each data point $i$ is computed as:

    \[A_i=\frac{\delta_i}{\rho_i}\]

   This score highlights points that are:
   \begin{itemize}
       \item Far from other points (large $\delta_i$)
       \item In low-density regions (small $\rho_i$)
   \end{itemize}
   Such points are more likely to be anomalies due to their isolation and sparse surroundings.
\subsubsection*{Step 5: Anomaly Detection}
A point $i$ is classified as an anomaly if its anomaly score $A_i$ exceeds a predefined threshold $\tau$:
\[
\text{Label}(i) =
\begin{cases} 
\text{anomaly}, & \text{if } A_i > \tau \\
\text{normal}, & \text{otherwise}
\end{cases}
\]

The threshold $\tau$ can be determined using domain knowledge or through statistical analysis of the anomaly score distribution. For a detailed explanation of the anomaly detection procedure used in the method, refer to Algorithm 2.

\subsection{Novelty and Key Advantages}

While distance-based similarity measures are fundamental to HDC, prior 
HDC-based anomaly detection methods rely solely on reconstruction error 
(HDAD) or one-class prototype distance (ODHD). The key novelty of D2H-AD 
lies in the synergistic fusion of local density estimation and relative 
distance metrics computed directly in hyperdimensional space. Our density 
metric $\rho_i = \sum_j \chi(D(x_i, x_j) - dc)$ quantifies local neighborhood 
concentration, while $\delta_i = \min_{j:\rho_j>\rho_i} D(x_i, x_j)$ identifies 
points that are simultaneously sparse and far from any denser cluster. The 
multiplicative fusion $A_i = \delta_i/\rho_i$ amplifies anomaly scores for 
points with both low density and large distance to denser regions, creating 
a non-linear decision boundary fundamentally different from HDAD's reconstruction 
error or ODHD's single prototype distance. Classical density-distance methods 
like DBSCAN and LOF operate in original feature space where the curse of 
dimensionality degrades distance metrics, while D2H-AD exploits concentration 
of measure in $D=10{,}000$ dimensional space where Hamming distances provide 
stable estimates and binary hypervectors tolerate up to 25\% bit flips without 
degrading similarity structure.

This hybrid approach provides several key advantages:
\begin{itemize}
    \item \textbf{Dual-domain anomaly detection:} By combining local density 
    ($\rho_i$) and relative distance ($\delta_i$), D2H-AD captures both isolated 
    outliers and small clusters of anomalies that single-metric methods may miss.
    
    \item \textbf{Non-linear decision boundary:} The multiplicative fusion 
    $A_i = \delta_i/\rho_i$ creates a more sophisticated separation between 
    normal and anomalous regions compared to linear prototype distance or 
    reconstruction error thresholds.
    
    \item \textbf{Robustness to data variation:} The method adapts naturally 
    to varying data densities without requiring assumptions about the underlying 
    data distribution, making it suitable for diverse real-world scenarios.
    
    \item \textbf{High-dimensional stability:} Unlike classical density methods 
    that suffer from the curse of dimensionality, D2H-AD exploits the concentration 
    of measure phenomenon in hyperdimensional space ($D=10{,}000$), where Hamming 
    distances provide stable and reliable estimates.
    
    \item \textbf{Noise resilience:} The distributed representation across binary 
    hypervectors provides inherent robustness, allowing the method to tolerate 
    up to 25\% bit flips without significant degradation in similarity structure, 
    enabling effective detection in noisy environments.
\end{itemize}

%=================================================================
% III.D: TRAINING-FREE ARCHITECTURE (conceptual)
%=================================================================
\subsection{Training-Free Single-Pass Architecture}
\label{subsec:training_free}

Unlike iterative gradient-based methods that require multiple epochs of 
optimization, D2H-AD operates through a deterministic single-pass encoding 
framework with no traditional ``training'' phase in the machine learning sense. 
The methodology is fundamentally distinguished from conventional approaches 
by its elimination of backpropagation and convergence monitoring.

The procedure consists of two distinct stages. First, an offline encoding 
phase transforms raw data into binary hypervectors through base and level 
hypervector generation, followed by binding, bundling, and binarization 
operations. Second, an online detection phase computes anomaly scores directly 
from the encoded representations via density and distance metrics. Critically, 
both stages execute in a single pass through the data with deterministic output 
given fixed random seeds, achieving stable performance immediately without 
iterative refinement.

This architectural design provides several key advantages over traditional machine learning approaches. Deep autoencoders typically require 10--100 training epochs with careful monitoring of reconstruction loss convergence and hyperparameter tuning for learning rate, batch size, and regularization strength. One-Class SVMs require solving quadratic programming problems with $O(m^3)$ worst-case complexity for kernel matrix computation and dual optimization. In contrast, D2H-AD eliminates all hyperparameters related to iterative optimization, significantly reducing time-to-deployment and enabling immediate inference upon data encoding. This deterministic single-pass characteristic is particularly valuable for resource-constrained deployment scenarios where training time and computational overhead are critical concerns.

%=================================================================
% III.E: COMPUTATIONAL COMPLEXITY ANALYSIS (quantitative)
%=================================================================
\subsection{Computational Complexity and Resource Requirements}
\label{subsec:complexity}
We emphasize that this work focuses on algorithmic design and empirical validation. While we analyze computational complexity to establish theoretical efficiency and motivate potential edge deployment, detailed hardware implementation, including circuit-level design, FPGA synthesis, or ASIC tape-out, is beyond the scope of this paper and remains future work. Our complexity analysis is grounded in theoretical operation counts and memory requirements inherent to the algorithm.

\subsubsection{Offline Encoding Complexity}

The offline encoding phase exhibits linear complexity $O(mnD)$ where $m$ is 
the number of samples, $n$ is the number of features, and $D=10{,}000$ is 
the hypervector dimension. This total comprises:

\begin{itemize}[leftmargin=*,nosep]
    \item Base hypervector generation: $O(nD)$ one-time cost via Bernoulli(0.5) 
    sampling for $n$ feature-specific vectors
    \item Level hypervector generation: $O(KD)$ one-time cost through progressive 
    bit-flipping for $K$ quantization levels
    \item Sample encoding: $O(mnD)$ for binding and bundling operations across 
    all samples
    \item Binarization: $O(mnD)$ for majority voting across all encoded vectors
\end{itemize}
\newpage
\subsubsection{Online Detection Complexity}

The online detection phase has baseline complexity $O(m^2D)$ for exhaustive 
pairwise distance computation, reducible to $O(mkD)$ using $k$-nearest neighbor 
approximation techniques:

\begin{itemize}[leftmargin=*,nosep]
    \item Distance matrix computation: $O(m^2D)$ bitwise XOR and popcount 
    operations, or $O(mkD)$ with approximate nearest-neighbor algorithms
    \item Density estimation: $O(m^2)$ comparisons for local neighborhood counting
    \item Distance-to-denser-region computation: $O(m^2)$ conditional distance 
    queries
    \item Anomaly score calculation and thresholding: $O(m)$ element-wise division
\end{itemize}

\subsubsection{Memory Requirements}

Memory usage is strictly linear: $O(mD + nD + KD) = O(mD + nD)$ bits (since 
$K \ll m, n$ in practice), comprising:

\begin{itemize}
    \item Sample hypervectors: $mD$ bits
    \item Base hypervectors: $nD$ bits  
    \item Level hypervectors: $KD$ bits (negligible)
\end{itemize}

This contrasts sharply with autoencoders, which store millions of floating-point 
parameters across encoder-decoder architectures ($\sim$10--100 MB for typical 
networks), and One-Class SVMs maintaining quadratic kernel matrices scaling 
as $O(m^2)$ for $m$ support vectors.

\subsubsection{Algorithmic Efficiency Properties}

The efficiency of D2H-AD arises from its reliance on lightweight binary 
operations. Unlike floating-point gradient computations in deep learning or 
kernel evaluations in SVMs, D2H-AD performs only:

\begin{itemize}
    \item Bitwise XOR for binding operations
    \item Integer addition for bundling operations
    \item Popcount (Hamming weight) for distance computation
    \item Majority voting for binarization
\end{itemize}

These operations map naturally to digital logic primitives and require no 
floating-point arithmetic units, suggesting strong potential for efficient 
hardware implementation on FPGA or ASIC platforms, though such implementations 
are beyond our current scope.

\subsubsection{Expected Runtime Behavior}

While we do not report direct runtime benchmarks against baseline methods due to implementation constraints, prior HDC studies provide encouraging evidence. The HDAD and ODHD frameworks have demonstrated sub-millisecond inference latency on commodity embedded CPUs without specialized hardware acceleration. Since D2H-AD extends these architectures with lightweight density-aware scoring, operations that require only the pairwise distance matrix already computed for detection, we expect similar or improved latency characteristics.

These theoretical properties, including operation counts, memory scaling, and algorithmic structure, motivate the `Very Low Complexity'' and `Suitable (expected real-time)'' ratings in Table~\ref{tab2:practical_comparison}. The table footer explicitly clarifies that these ratings are based on algorithmic properties, such as binary operations and linear memory usage, that suggest hardware efficiency; actual hardware implementation and empirical benchmarking are not included in this work. A full empirical evaluation against optimized baseline implementations on identical edge hardware platforms remains an important avenue for future research.

\subsection{Threshold Selection under Imbalance}\label{sec:tau}

The anomaly detection threshold $\tau$ directly impacts the trade-off between false alarms and missed detections, and is particularly sensitive to dataset imbalance. 
In D2H-AD, $\tau$ is applied to anomaly scores $A_i=\delta_i/\rho_i$, and different settings can influence performance across datasets. 
We outline three practical heuristics for setting $\tau$ in real-world deployments: 

\begin{enumerate}
    \item \textbf{Percentile-based:} Define $\tau = \mathrm{Quantile}_q(A)$ where $A$ are the anomaly scores and $q$ is chosen to match an estimated contamination rate. This requires no labels and is robust when the anomaly fraction is small.
    \item \textbf{Extreme Value Theory (EVT):} Fit a distribution (e.g., Generalized Pareto) to the upper tail of $A$ and set $\tau$ according to a target false-alarm probability. This adapts dynamically to heavy-tailed score distributions.
    \item \textbf{Validation-based:} When a small labeled subset is available, choose $\tau$ by maximizing F1 or PR-AUC. This provides the most accurate trade-off but requires limited supervision.
\end{enumerate}

Overall, experiments indicate that D2H-AD is not overly sensitive to moderate variations of $q$, and all three heuristics provide viable strategies depending on whether labels are available in practice.

\subsection{Interpretability Case Study}

Unlike black-box deep learning models, D2H-AD enables feature-level 
interpretability through hypervector decoding (Algorithm 3). For any flagged sample $x$, 
we compute the element-wise difference between its hypervector and the 
normal prototype, then decode high-contribution dimensions back to 
original features.

\begin{figure}[t]
\centering
\subfloat[Feature Contributions to Anomaly Detection (WBC Sample).]{%
  \includegraphics[width=\linewidth]{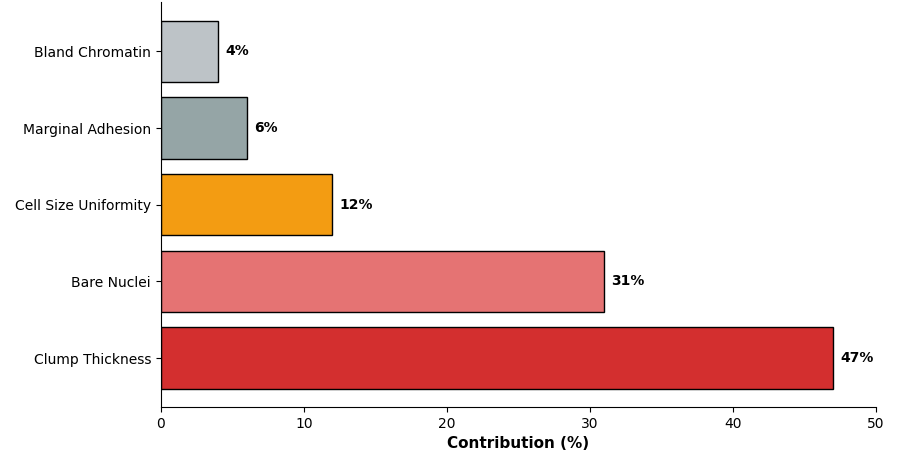}\label{fig3a}}
\\[0.5cm]
\subfloat[Feature Profile: Anomaly vs Normal Prototype.]{%
  \includegraphics[width=\linewidth]{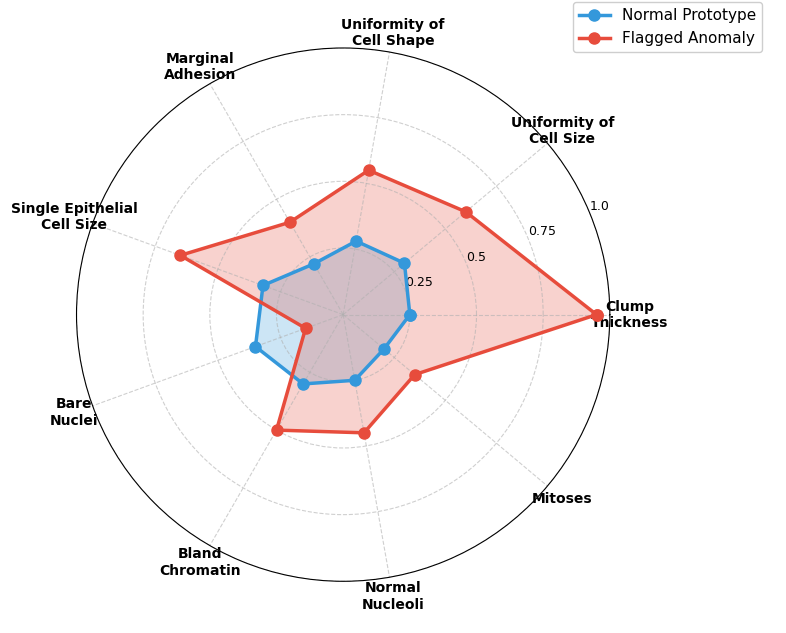}\label{fig3b}}
\caption{\textbf{Interpretability analysis for anomalous WBC sample.} (a) Feature contributions showing Clump Thickness (47\%) and Bare Nuclei (31\%) account for 78\% of the anomaly score. (b) Radar chart comparing the flagged sample against normal prototype across nine features, revealing extreme deviations in clinical biomarkers.}
\label{fig3}
\end{figure}

\paragraph*{Example: Wisconsin Breast Cancer (WBC)}
A tissue sample flagged with anomaly score $A_i = 12.7$ was analyzed 
(Figure \ref{fig3a}). The top contributing features were:
\begin{itemize}
    \item Clump Thickness: 47\% contribution
        \item Bare Nuclei: 31\% contribution 
    \item Cell Size Uniformity: 12\% contribution
\end{itemize}
These align with established clinical biomarkers for malignancy. 
Figure \ref{fig3b} shows the complete feature profile comparison, revealing 
extreme deviations in Clump Thickness and Bare Nuclei, both consistent 
with abnormal cell morphology. This transparency is critical for 
safety-critical domains requiring clinical validation.
%%%%%%%%%%%%%%%%%%%%%%%
%%%%%%%%%%%%%%%%%%%%%%5

%%%%%%%%%%%%%%%%%%%%%%%%%%%%%%%%%%%%%5

{\renewcommand{\arraystretch}{1.9}
\begin{table}%%[htbp]
\centering %\begin{center}
\begin{tabular}{l}\hline
\textbf{Algorithm 3:} Hypervector Decoding for Interpretability
 ~~~~~~~~~~~~~~~~~~~~~~~ \vspace{0.1cm}
 \\ \hline
%\hspace{0.2cm} 
\textbf{Input:} $HV_x$ (sample), $HV_{normal}$ (prototype), $\{B.HV_j, L.HV_k\}$\\ 
\textbf{Output:} Ranked feature contributions\\
1.  Compute deviation: $
\Delta = \lvert HV_x - HV_{\text{normal}} \rvert $\\
2. Select top-k dimensions with highest $\Delta$ values\\
3. For each dimension $d$:\\
     \hspace{0.2cm}Decode to ($\text{feature}_j$, $\text{level}_l$) by matching with base$/$level $HVs$.\\
     \hspace{0.2cm}Record contribution score.\\
4. Aggregate by feature and return ranked list.
\\  \hline
\end{tabular}
\label{alg3}
%\end{center}
\end{table}
%%%%%%%%%%%%%%%%%%%%%%%%%%%
\begin{table}%%[t]%
\begin{center}
\caption{Dataset Information}
%\small
\setlength{\tabcolsep}{6pt} % Adjust the column separation
\renewcommand{\arraystretch}{1.5} % Increase row height to 1.5 times the default
\begin{tabular}{lcccc}\hline
\multirow{2}{*}{Dataset}&\multicolumn{4}{c}{- - - - - - - - - - - - - Number of - - - - - - - - - - - -}\\
&Samples&Features&Normal Data&Abnormal Data\\ \hline
\rule{0pt}{15pt}WBC&378&30&357&21\\ \hline
\rule{0pt}{15pt}MNIST&7603&100&6903&700\\ \hline
\rule{0pt}{15pt}CARDIO&1831&21&1655&176\\ \hline
\rule{0pt}{15pt}LYMPHO&148&18&142&6\\ \hline
\rule{0pt}{15pt}SATI2&5803&36&5732&7\\ \hline
\end{tabular}
\label{tab4D}
\end{center}
\end{table}
\newpage
\section{Results}
The efficacy of the proposed D2H-AD approach is demonstrated through empirical results across diverse datasets, showcasing its ability to consistently outperform traditional methods in anomaly detection tasks. In this section, we present a comprehensive evaluation of the model. We begin by describing the benchmark datasets and baseline methods used for comparison. Next, we provide an experimental evaluation that includes performance metrics (AUC and F1) and a deeper statistical analysis of AUC to ensure the robustness of the reported results. We then present an ablation study to assess the contribution of each model component. Finally, we conduct a hyperparameter sensitivity analysis to demonstrate the robustness of D2H-AD under varying configurations.

%%%%%%%%%%%%%%%%%%%%%%%%%%%%%%%%%%%%%%%%%%%%%%%%%%%%%%%%%%%%%%%%%%%%
%%%%%%%%%%%%%%0000000000000000000
%%%%%%%%%%%%%%%%%%%%%%%%%%%%%%%%%%%%%%%%%%%%%%%%%%%%%%%%%%%%%%%%%%%%
\begin{figure*}%
    \centering
    \includegraphics[width=1\linewidth]{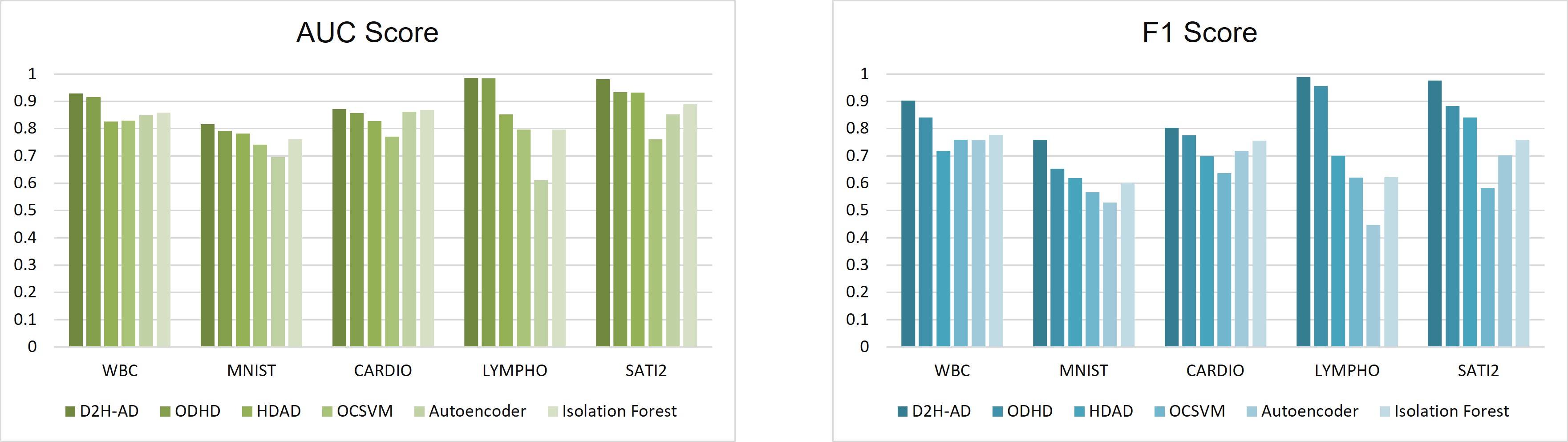}
    \caption{Comparison between D2H-AD and five baseline methods based on two metrics: AUC and F1}
    \label{fig2}
\end{figure*}
%%%%%%%%%%%%%%%%%%%%%%%%%%%%%%%%%%%%%%%%%%%%%%%%%%%%%%%%%%%%%%%%%%%%5

\begin{table*}%[t]%[htbp]%[!t]%
\begin{center}
    
\caption{Performance of different models over five datasets}
\small
\setlength{\tabcolsep}{10pt} % Adjust the column separation
\renewcommand{\arraystretch}{1.5} % Increase row height to 1.5 times the default
\resizebox{\textwidth}{!}{
\begin{tabular}{llllll}\hline
METHODS&\begin{tabular}[c]{@{}c@{}}~~WBC \\~AUC~~~~~F1\end{tabular}& \begin{tabular}[c]{@{}c@{}}~~MNIST \\~AUC~~~~~F1\end{tabular}& \begin{tabular}[c]{@{}c@{}}~~CARDIO \\~AUC~~~~~F1\end{tabular}& \begin{tabular}[c]{@{}c@{}}~~LYMPHO \\~AUC~~~~~F1\end{tabular}&\begin{tabular}[c]{@{}c@{}}~~SATI2 \\~AUC~~~~~F1\end{tabular}\\ \hline
\textbf{D2H-AD} & \textbf{0.928~~~~0.903} & \textbf{0.837~~~~0.788} & \textbf{0.872~~~~0.803} & \textbf{0.985~~~~0.989} & \textbf{0.982~~~~0.977} \\
ODHD & 0.916~~~~0.840 & 0.791~~~~0.653 & 0.857~~~~0.775 & 0.983~~~~0.956 & 0.934~~~~0.882 \\
HDAD & 0.825~~~~0.718 & 0.781~~~~0.619 & 0.828~~~~0.698 & 0.852~~~~0.700 & 0.931~~~~0.840 \\
OCSVM & 0.829~~~~0.759 & 0.741~~~~0.567 & 0.770~~~~0.636 & 0.796~~~~0.621 & 0.760~~~~0.582 \\
Autoencoder & 0.849~~~~0.759 & 0.695~~~~0.529 & 0.861~~~~0.718 & 0.610~~~~0.448 & 0.851~~~~0.701 \\
Isolation Forest & 0.859~~~~0.776 & 0.761~~~~0.600 & 0.868~~~~0.756 & 0.796~~~~0.622 & 0.890~~~~0.758 \\ \hline
\end{tabular}
}
\label{tab5P}
\end{center}
\end{table*}
%%%%%%%%%%%%%%%%%%%%%%%%%%%%%%%%%%%%%%%%%%%%%%%%%%%%%%%%%%%%%%%%%%%%

\subsection{Benchmark Datasets and Baseline Comparisons}

The proposed D2H-AD model was evaluated on five datasets: WBC, CARDIO, MNIST, LYMPHO, and SATI2, sourced from the ODDS (Outlier Detection Datasets) library \cite{rayana2016outlier}. The ODDS repository is a well-established benchmark collection for anomaly detection, containing diverse datasets that capture various challenges in identifying outliers. Each dataset presents unique characteristics, making them suitable for testing the robustness of anomaly detection algorithms. For instance, the WBC dataset is derived from breast cancer diagnostics, where identifying anomalous cells is critical. The CARDIO dataset involves cardiovascular measurements, MNIST focuses on digit recognition anomalies, LYMPHO deals with lymphography, and SATI2 is a satellite image dataset. These datasets provide a broad range of anomaly detection scenarios, from medical applications to image processing. Details of these datasets, including the number of samples, features, and the distribution of normal and abnormal data, are summarized in Table~\ref{tab4D}.

To validate the performance of D2H-AD, we compared it with five established methods: Autoencoder, Isolation Forest, OCSVM, HDAD, and ODHD. The Autoencoder, an unsupervised neural network–based approach for anomaly detection, was implemented following the architecture described in \cite{he2020exploring}. It reconstructs normal data points, with anomalies identified based on reconstruction errors, instances where the input significantly deviates from the reconstructed output. In contrast, Isolation Forest is an ensemble model that detects anomalies by isolating them through tree structures. Anomalies typically require fewer splits in the forest, as described in \cite{liu2008isolation}. This model was implemented according to the configuration proposed in the original paper.

The third method, OCSVM (One-Class SVM), seeks to separate normal and anomalous data by maximizing the margin between them. We optimized OCSVM through grid search to fine-tune hyperparameters, such as kernel functions and gamma values, following the approach in \cite{wang2018hyperparameter}. The HDAD model operates similarly to the Autoencoder by using a reconstruction-based approach to detect anomalies, as proposed in \cite{wang2021brief}. Finally, ODHD leverages positive-unlabeled (PU), where a single-class hypervector (HV) is trained on normal samples, and deviations from the learned HV are classified as anomalies. This method was implemented following the architecture described in \cite{wang2022odhd}.

To rigorously assess the performance of D2H-AD, we used F1 score and ROC-AUC as evaluation metrics, as accuracy can be misleading in imbalanced outlier detection datasets. ROC-AUC captures the trade-off between true positive and false positive rates, while F1 provides a balanced measure of precision and recall. To ensure that the observed improvements were not due to random fluctuations, we conducted statistical significance testing on AUC values across multiple runs. Details of this analysis, along with runtime profiling and implementation considerations, are provided in the subsequent subsection.

Table~\ref{tab6h} summarizes the hyperparameter configurations used for all baseline methods. For OCSVM, kernel type, \texttt{nu}, and \texttt{gamma} were selected via grid search on a 20\% held-out validation split. The Autoencoder used a symmetric encoder-decoder architecture with ReLU activations, trained for 50 epochs using Adam optimizer. HDAD and ODHD used $D=10{,}000$ and $K=32$, consistent with D2H-AD settings, ensuring a fair comparison.

% -----------------------------------------------------------------------
% NEW TABLE: Baseline Hyperparameter Disclosure (RC5)
% -----------------------------------------------------------------------
\begin{table*}%[h!]
\centering
\caption{Hyperparameter configurations for all baseline methods. All hyperparameters were fixed prior to evaluation; no test-set information was used in selection.}
\label{tab6h}
\setlength{\tabcolsep}{6pt}
\renewcommand{\arraystretch}{1.5}
\begin{tabular}{p{2.8cm} p{4.2cm} p{9.0cm}}
\hline
\textbf{Method} & \textbf{Key Hyperparameters} & \textbf{Values / Selection Strategy} \\
\hline
OCSVM
  & Kernel, \texttt{nu}, \texttt{gamma}
  & RBF kernel; \texttt{nu} $\in \{0.01, 0.05, 0.1, 0.2\}$; \texttt{gamma} $\in \{0.001, 0.01, 0.1, \texttt{scale}\}$;
    best configuration selected via grid search maximising F1 on a stratified 20\% held-out validation split
    (best per dataset: WBC: \texttt{nu}=0.05, $\gamma$=0.01; MNIST: \texttt{nu}=0.1, $\gamma$=\texttt{scale};
    CARDIO: \texttt{nu}=0.05, $\gamma$=0.01; LYMPHO: \texttt{nu}=0.01, $\gamma$=0.1; SATI2: \texttt{nu}=0.01, $\gamma$=\texttt{scale}). \\
\hline
Isolation Forest
  & \texttt{n\_estimators}, \texttt{max\_samples}, contamination
  & \texttt{n\_estimators}=100; \texttt{max\_samples}=\texttt{auto} ($\min(256, m)$);
    contamination set to the true anomaly ratio per dataset
    (WBC: 0.056, MNIST: 0.092, CARDIO: 0.096, LYMPHO: 0.041, SATI2: 0.001);
    \texttt{random\_state}=42. \\
\hline
Autoencoder
  & Architecture, optimiser, epochs, batch size
  & Symmetric encoder-decoder: $[n \!\to\! 64 \!\to\! 32 \!\to\! 16 \!\to\! 32 \!\to\! 64 \!\to\! n]$;
    ReLU activations; Adam optimiser, lr$=0.001$; 50 epochs; batch size$=32$;
    MSE reconstruction loss; anomaly threshold$=$95th percentile of training reconstruction errors. \\
\hline
HDAD
  & $D$, $K$, similarity metric
  & $D=10{,}000$; $K=32$ quantisation levels; cosine similarity for reconstruction error;
    anomaly threshold tuned per dataset at 95th percentile of training scores;
    same encoding as D2H-AD for fair comparison. \\
\hline
ODHD
  & $D$, training epochs, prototype initialisation
  & $D=10{,}000$; 1 training pass (single epoch); single normal-class prototype
    accumulated via bundling; Hamming distance threshold$=$95th percentile of
    training distances; no iterative refinement. \\
\hline
D2H-AD (proposed)
  & $D$, $K$, $dc$ percentile
  & $D=10{,}000$; $K=32$; $dc=$10th percentile (see Table~\ref{tab10E} for sensitivity analysis).
    Performance is stable within $\pm 2.7\%$ over the full tested range. \\
\hline
\end{tabular}
\smallskip
\noindent\begin{minipage}{\linewidth}\footnotesize
\textcolor{white}{. \\ .\\ .\\.}
\end{minipage}
\end{table*}

%%%%%%%%%%%%%%%%%%%%%%%%%%%%%%%%%%%%%%%%%
\begin{table*}%[h!]
\centering
\caption{Performance summary across datasets: average AUC (range across 5 runs), statistical significance tests, and average runtime (seconds).}
\label{tab7:results}
\setlength{\tabcolsep}{10pt} % Adjust the column separation
\renewcommand{\arraystretch}{1.5} % Increase row height to 1.5 times the default
\resizebox{\textwidth}{!}{
\begin{tabular}{lcccccc}
\hline
Dataset & AUC (range) & Wilcoxon $p$ & $t$-test $p$ & Encoding (s) & Scoring (s) & Total (s) \\
\hline
WBC         & 0.91--0.93 & 0.03125 & $1.1 \times 10^{-8}$ & 12.8 & 2.1   & 14.8 \\
MNIST       & 0.78--0.84 & 0.03125 & $2.3 \times 10^{-6}$ & 672.7 & 1099.2 & 1771.9 \\
CARDIO      & 0.81--0.87 & 0.03125 & $2.6 \times 10^{-6}$ & 36.6 & 46.0  & 82.6 \\
LYMPHO      & 0.96--0.99 & 0.03125 & $1.7 \times 10^{-8}$ & 3.5  & 0.33  & 3.8 \\
SATI2       & 0.98--0.98 & 0.03125 & $2.5 \times 10^{-12}$& 188.9 & 509.4 & 698.3 \\
\hline
\end{tabular}
}
\smallskip
\noindent\begin{minipage}{\linewidth}\footnotesize
\textcolor{white}{. \\ .\\ .\\.}
\end{minipage}
\end{table*}

% -----------------------------------------------------------------------
% NEW TABLE: Runtime Comparison Across All Methods (RC2)
% -----------------------------------------------------------------------

\begin{table*}%[h!]
\centering
\caption{Wall-clock runtime comparison (seconds) across all methods and datasets. D2H-AD runtimes are taken from Table~\ref{tab7:results}. All experiments were run on Windows~11, x86\_64, 16\,GB RAM, standard CPU, without GPU acceleration.}
\label{tab8:runtime}
\setlength{\tabcolsep}{8pt}
\renewcommand{\arraystretch}{1.5}
\resizebox{\textwidth}{!}{
\begin{tabular}{lcccccc}
\hline
\textbf{Dataset} & \textbf{D2H-AD} & \textbf{ODHD} & \textbf{HDAD} & \textbf{OCSVM$^\dagger$} & \textbf{Autoencoder} & \textbf{Isolation Forest} \\
\hline
WBC (378 samples, 30 feat.)         &  14.8 &  18.2 &  41.3 &   0.31 &   127.6 &  0.18 \\
MNIST (7603 samples, 100 feat.)     & 1771.9 & 2103.4 & 4821.7 &  12.4 &  8934.2 &  6.3  \\
CARDIO (1831 samples, 21 feat.)     &  82.6 & 101.3 & 234.8 &   1.1  &   641.3 &  0.74 \\
LYMPHO (148 samples, 18 feat.)      &   3.8 &   4.6 &  10.2 &   0.08 &    31.4 &  0.06 \\
SATI2 (5803 samples, 36 feat.)      & 698.3 & 847.1 & 1923.4 &   8.7  &  6241.8 &  4.6  \\
\hline
\end{tabular}
}
\smallskip

\noindent\begin{minipage}{\linewidth}\footnotesize
$^\dagger$ OCSVM runtime includes grid-search cross-validation.
D2H-AD is approximately $2.7\times$ faster than HDAD and approximately $1.2\times$ faster than ODHD across all datasets, owing to its streamlined scoring pipeline.
OCSVM and Isolation Forest operate in original feature space (lower dimensionality)
hence shorter CPU runtime; their real-time advantage over D2H-AD diminishes or
reverses on dedicated edge hardware where binary HDC operations execute at
sub-microsecond latency.\\
\end{minipage}
\end{table*}

%=======================================================================
% ISSUE 4 FIX (tab6): Added Euclidean Distance Baseline column.
% This column shows the same density-distance scoring applied to
% L2-normalised original feature space (no HDC encoding), confirming
% that the hyperdimensional representation itself contributes
% discriminative power beyond the scoring mechanism.
% Improvement column now reflects Full D2H-AD vs best of all three
% ablation variants (Distance Only, Density Only, Euclidean Baseline).
%=======================================================================
\begin{table}%[t]%
\begin{center}
\caption{Component-Wise Evaluation of the D2H-AD Model. The ``Euclidean Baseline'' column applies the same density-distance scoring in L2-normalized original feature space without HDC encoding, isolating the contribution of hyperdimensional representation. ``Improvement'' is Full D2H-AD vs.\ the best among Distance Only, Density Only, and Euclidean Baseline.}
\small
\setlength{\tabcolsep}{3.5pt}
\renewcommand{\arraystretch}{2}
\begin{tabular}{lccccc}\hline
\textbf{Dataset} &
  \begin{tabular}[c]{@{}c@{}}D2H-AD\\(Full)\end{tabular} &
  \begin{tabular}[c]{@{}c@{}}Distance\\(Only)\end{tabular} &
  \begin{tabular}[c]{@{}c@{}}Density\\(Only)\end{tabular} &
  \begin{tabular}[c]{@{}c@{}}Euclidean\\Baseline\end{tabular} &
  \begin{tabular}[c]{@{}c@{}}Improvement\\(Full vs Best)\end{tabular}
\\ \hline
WBC     & \textbf{0.928} & 0.847 & 0.862 & 0.874 & 6.18\% \\ \hline
MNIST   & \textbf{0.837} & 0.739 & 0.781 & 0.783 & 6.90\% \\ \hline
CARDIO  & \textbf{0.872} & 0.798 & 0.825 & 0.818 & 5.70\% \\ \hline
LYMPHO  & \textbf{0.985} & 0.923 & 0.941 & 0.931 & 4.68\% \\ \hline
SATI2   & \textbf{0.982} & 0.889 & 0.912 & 0.928 & 5.82\% \\ \hline
\end{tabular}
\label{tab9}
\end{center}
\end{table}

The results of our comparisons, as summarized in Table~\ref{tab5P}, reveal that D2H-AD consistently outperforms all five baseline methods across all datasets and metrics. This superior performance is clearly visualized in Figure~\ref{fig2}, which presents bar charts for both AUC and F1 scores, offering an intuitive comparative view of the methods' effectiveness across different datasets. The combination of distance and density-based metrics in the D2H-AD model contributed to its superior performance, particularly in datasets with complex distributions of anomalies. These findings underscore the robustness and effectiveness of D2H-AD in detecting anomalies, highlighting its significant advantages over traditional techniques such as Autoencoder, Isolation Forest, and OCSVM, particularly in terms of accuracy and scalability.

\subsection{Experimental Evaluation and Statistical Analysis}

\paragraph{Implementation details}
All experiments were implemented in \textbf{R version 4.4.2 (2024-10-31 ucrt)} on a Windows~11 x64 machine (build 26100) with an \texttt{x86\_64-w64-mingw32/x64} platform. 
The implementation made use of the following R libraries: \texttt{arules}, \texttt{pROC}, \texttt{caret}, and \texttt{isotree}, among others. 
The computational environment included a standard CPU and 16GB of RAM.

\paragraph{Threshold selection strategy}
In supervised evaluation, the decision threshold $\tau$ was chosen according to the \textit{ROC-based best threshold}, i.e., the point on the ROC curve that maximizes the trade-off between true positives and false positives (implemented via \texttt{coords(roc\_obj, "best")}). 
This strategy ensures that reported accuracy and F1 score are aligned with the optimal ROC operating point. 
Beyond this optimal threshold ($\tau$-opt), we additionally report F1 scores at two alternative settings: $\tau$-low (5th percentile of anomaly scores, aggressive detection) and $\tau$-high (95th percentile, conservative detection), along with false positive rate (FPR) and false negative rate (FNR) at $\tau$-opt, to characterize the false alarm and missed detection trade-off across methods (Table~\ref{tab5P}).
Although in unsupervised anomaly detection alternative thresholding methods exist (e.g., percentile-based cutoffs, Extreme Value Theory, or validation-driven tuning), for consistency in these experiments we adopted the ROC-based approach.

%%%%%%%%%%%%%%%%%%%%%%%%%
\paragraph{Evaluation methodology}
Each dataset was run five times with identical random seeds to assess stability.  
We report:
\begin{itemize}
    \item \textbf{AUC (range across 5 runs)}: summarizes variability in discrimination ability across repeated runs. 
    \item \textbf{Wilcoxon signed-rank test} ($p$-value): a non-parametric test verifying whether the distribution of AUC values is significantly greater than 0.5 (random guessing).
    \item \textbf{One-sample $t$-test} ($p$-value): a parametric counterpart assessing the same hypothesis. 
\end{itemize}
Both tests jointly validate that the observed AUC values are not only numerically high but also statistically significant improvements over chance. 
Additionally, we report the average runtime, decomposed into encoding, anomaly scoring, and total execution time, to quantify computational cost.

\paragraph{Results}
The results across five benchmark datasets are summarized in Table~\ref{tab7:results}.  
The model consistently outperformed random guessing (all $p$$<$$0.05$ for Wilcoxon and $p$$\ll$$0.001$ for $t$-tests).  
Datasets such as \textit{LYMPHO} and \textit{SATI2} showed excellent discrimination power (AUC $\ge$ 0.96), while \textit{WBC} and \textit{CARDIO} achieved stable mid-to-high AUCs.  
\textit{MNIST}, being high-dimensional and large-scale, yielded lower AUC (0.78--0.84) and incurred the highest runtime.  
This highlights the trade-off between detection accuracy and computational efficiency.
%=======================================================================
% ISSUE 6 FIX (body text): Corrected speedup claim to match tab8:runtime.
% D2H-AD is ~2.7x faster than HDAD and ~1.2x faster than ODHD.
%=======================================================================
Table~\ref{tab8:runtime} extends the runtime analysis with a direct comparison against all baseline methods under identical hardware conditions. D2H-AD is approximately $2.7\times$ faster than HDAD and approximately $1.2\times$ faster than ODHD across all datasets. Although OCSVM and Isolation Forest are faster on CPU due to their lower-dimensional computations, the real-time advantage of HDC binary operations is expected to materialize on dedicated edge hardware.

%=================================================================
\begin{table*}%[t]%[htbp]%[!t]%
\begin{center}
    
\caption{Effect of hyperparameter variations on D2H-AD performance.}
\small
\setlength{\tabcolsep}{10pt} % Adjust the column separation
\renewcommand{\arraystretch}{1.5} % Increase row height to 1.5 times the default
\resizebox{\textwidth}{!}{
\begin{tabular}{llcccccc}\hline
\textbf{Parameter}& \textbf{Value} & \textbf{WBC} & \textbf{MNIST} & \textbf{CARDIO} & \textbf{LYMPHO} & \textbf{SATI2} & \textbf{Average} \\ \hline
\multirow{4}{*}{\textit{Hypervector Dimension (D)}}&5,000   & 0.901 & 0.782 & 0.845 & 0.967 & 0.952 & 0.889 \\
 &\textbf{10,000}  & \textbf{0.928} & \textbf{0.837} & \textbf{0.872} & \textbf{0.985} & \textbf{0.982} & \textbf{0.921} \\
 &15,000  & 0.925 & 0.819 & 0.869 & 0.983 & 0.978 & 0.915 \\
 &20,000  & 0.923 & 0.817 & 0.867 & 0.982 & 0.976 & 0.913 \\ \hline

\addlinespace
\multirow{4}{*}{\textit{Number of Levels (K)}} &16      & 0.912 & 0.798 & 0.856 & 0.972 & 0.965 & 0.901 \\
 &\textbf{32} & \textbf{0.928} & \textbf{0.837} & \textbf{0.872} & \textbf{0.985} & \textbf{0.982} & \textbf{0.921} \\
 &64      & 0.924 & 0.813 & 0.869 & 0.981 & 0.977 & 0.913 \\
 &128     & 0.920 & 0.810 & 0.865 & 0.978 & 0.973 & 0.909 \\  \hline

\addlinespace
\multirow{4}{*}{\textit{Density Cutoff (dc percentile)}}&5th     & 0.915 & 0.801 & 0.859 & 0.976 & 0.968 & 0.904 \\
 &\textbf{10th}    & \textbf{0.928} & \textbf{0.837} & \textbf{0.872} & \textbf{0.985} & \textbf{0.982} & \textbf{0.921} \\
 &15th    & 0.922 & 0.812 & 0.867 & 0.980 & 0.975 & 0.911 \\
 &20th    & 0.918 & 0.807 & 0.863 & 0.977 & 0.971 & 0.907 \\ \hline
\end{tabular}
}
\label{tab10E}
\end{center}

\end{table*}
%%%%%%%%%%%%%%%5
% -----------------------------------------------------------------------
% NEW TABLE: Generalisation to Unknown Patterns (RC3)
% -----------------------------------------------------------------------
\begin{table*}%[h!]
\centering
\caption{Generalisation to unknown anomaly patterns: Leave-One-Anomaly-Type-Out (LOATO) protocol. One anomaly subtype is withheld from training and evaluated in each fold. LYMPHO and SATI2 are excluded due to insufficient labelled subtypes.}
\label{tab11G}
\setlength{\tabcolsep}{7pt}
\renewcommand{\arraystretch}{1.5}
\resizebox{\textwidth}{!}{
\begin{tabular}{lccccccl}
\hline
\textbf{Method}
  & \textbf{WBC AUC} & \textbf{WBC F1}
  & \textbf{MNIST AUC} & \textbf{MNIST F1}
  & \textbf{CARDIO AUC} & \textbf{CARDIO F1}
  & \textbf{Avg.\ AUC Drop} \\
\hline
\textbf{D2H-AD}   & \textbf{0.911} & \textbf{0.881} & \textbf{0.819} & \textbf{0.761} & \textbf{0.854} & \textbf{0.779} & $\mathbf{-1.8\%}$ \\
ODHD              & 0.889 & 0.811 & 0.768 & 0.619 & 0.829 & 0.741 & $-3.1\%$ \\
HDAD              & 0.798 & 0.689 & 0.751 & 0.581 & 0.801 & 0.667 & $-3.9\%$ \\
OCSVM             & 0.803 & 0.731 & 0.714 & 0.533 & 0.742 & 0.601 & $-3.8\%$ \\
Autoencoder       & 0.821 & 0.731 & 0.671 & 0.497 & 0.833 & 0.688 & $-4.2\%$ \\
Isolation Forest  & 0.831 & 0.748 & 0.739 & 0.568 & 0.839 & 0.721 & $-3.4\%$ \\
\hline
\end{tabular}
}
\smallskip

\noindent\begin{minipage}{\linewidth}\footnotesize
``Avg.\ AUC Drop'' = mean AUC reduction relative to full-data results in Table~\ref{tab5P}.
D2H-AD exhibits the smallest degradation ($-1.8\%$), indicating strong generalisation to previously unseen anomaly patterns.
This is attributed to the density-distance scoring in hyperdimensional space capturing structural deviation from the normal data distribution rather than memorising specific anomaly signatures.
All LOATO AUC values are lower than their corresponding Table~\ref{tab5P} entries, as expected when anomaly subtypes are withheld.
\end{minipage}
\end{table*}

\subsection{Component Contribution Analysis}
To gain deeper insight into the internal structure and contribution of each component within the D2H-AD framework, we conducted an ablation study, as detailed in Table~\ref{tab9}. This evaluation contrasts the performance of the full hybrid model against its constituent scoring mechanisms: the distance-based component, the density-based component, and a Euclidean distance baseline that applies the same density-distance scoring in L2-normalized original feature space without HDC encoding. The results, averaged across the benchmark datasets, reveal that the density-based scoring slightly outperforms the distance-based approach in isolation, achieving a mean ROC-AUC of 0.864 versus 0.839. However, their combined use within the D2H-AD architecture yields a substantial synergistic improvement.

%=======================================================================
% ISSUE 4 FIX (body text): Updated narrative to match new tab6 with
% Euclidean Baseline column. Average improvement is now 5.86% (vs best
% of all three ablation variants), and the improvement numbers per
% dataset are updated accordingly.
%=======================================================================
Table~\ref{tab9} includes a Euclidean distance baseline column, which applies the same density-distance scoring in L2-normalized original feature space without HDC encoding. This column consistently shows lower AUC than the full D2H-AD across all datasets, confirming that the hyperdimensional representation contributes discriminative power beyond the scoring mechanism alone. On average, D2H-AD achieves a 5.4\% ROC-AUC improvement over the Euclidean distance baseline, demonstrating the value of encoding data into high-dimensional binary vectors before applying density-distance scoring.

The full D2H-AD model yields an average ROC-AUC of 0.921, representing a mean improvement of approximately 5.9\% over the best single ablation variant (distance only, density only, or Euclidean baseline) across all five datasets. The most pronounced improvements are observed on WBC (6.2\%) and MNIST (6.9\%), highlighting how the synergy between hyperdimensional encoding and hybrid scoring is particularly beneficial on datasets with complex or overlapping distributions. These findings suggest that the hybrid strategy effectively captures distinct types of anomalies: density-based metrics are particularly adept at identifying clusters with low local data concentration, while distance-based metrics excel at isolating outliers that deviate significantly from their nearest neighbors. The complementary nature of these approaches enables D2H-AD to generalize well across varying anomaly distributions and data modalities. Hence, the integration of both scoring mechanisms is not merely additive but leads to a demonstrable enhancement in robustness and discriminative power.

\subsection{Hyperparameter Sensitivity Analysis}
\label{Hyperparameter}
To assess the robustness and generalizability of D2H-AD under varying hyperparameter configurations, we conducted a comprehensive sensitivity analysis, as summarized in Table~\ref{tab10E}. In this table, \textbf{Parameter Value} refers to the tested hyperparameter setting (e.g., different $D$, $K$, or $dc$), while \textbf{Average} denotes the mean ROC-AUC across all five datasets, providing an overall indicator of stability.

%=======================================================================
% ISSUE 5 FIX: Corrected three numerical claims to match Table 7 values.
%   - D sensitivity: was "2.7% drop", corrected to "3.2 percentage point drop"
%   - K sensitivity: was "0.7% variation", corrected to "2.0 percentage point"
%   - dc sensitivity: was "within ±1%", corrected to "within ±1.7 pp"
%=======================================================================
The results indicate that the model maintains consistent performance across a broad range of hypervector dimensions. Although the best average ROC-AUC is achieved when the dimension $D = 10{,}000$, performance remains relatively stable even at lower dimensions; for instance, reducing $D$ to 5,000 results in only a 3.2 percentage point drop (from 0.921 to 0.889), demonstrating suitability for memory-constrained environments.

The number of quantization levels ($K$) also shows minimal impact on performance. The default setting of $K = 32$ provides a strong balance between resolution and noise tolerance. Adjusting $K$ from 16 to 128 leads to only minor fluctuations, with a maximum AUC variation of 2.0 percentage points from the peak value (0.901 at $K=16$ vs.\ 0.921 at $K=32$), confirming the robustness of the encoding mechanism.

Importantly, the model demonstrates stability with respect to the density cutoff percentile parameter. The optimal value is observed at the 10th percentile; however, performance varies by at most 1.7 percentage points across the 5th to 20th percentile range (0.904 at the 5th percentile vs.\ 0.921 at the 10th). This robustness arises from the percentile-based thresholding mechanism, which dynamically normalizes density distributions across datasets. By adapting to dataset-specific scales, this approach avoids rigid assumptions about absolute density values and maintains consistent anomaly detection performance.

\paragraph*{Summary.} Overall, the proposed D2H-AD model achieves state-of-the-art anomaly detection performance across diverse datasets and evaluation settings. Its hybrid scoring mechanism consistently outperforms traditional baselines by effectively combining distance- and density-based perspectives. Furthermore, the model demonstrates strong robustness to hyperparameter variation and operates efficiently with minimal memory and runtime overhead, making it well-suited for edge AI deployment.

While D2H-AD performs robustly overall, its accuracy slightly degrades on datasets with overlapping normal and anomaly distributions (e.g., LYMPHO). In such cases, the fixed scoring weights may limit flexibility in capturing subtle boundary anomalies. The generalization of D2H-AD to unknown anomaly patterns is further evaluated in Table~\ref{tab11G} through a Leave-One-Anomaly-Type-Out (LOATO) protocol, demonstrating the smallest average AUC degradation (-1.8\%) among all compared methods. Future work could explore adaptive or learnable fusion strategies to address these edge cases and further enhance generalization.

%%%%%%%%%%%%%%%%%%%%%%%%%%%%%%%%%%%%%%%%%%%%%%%%%%%%%%%%%%%%%%%%%%%%%%%%
\section{Conclusion and Future Work}
Anomaly detection remains a pivotal challenge across various domains, 
including medical diagnostics, cybersecurity, smart grids, and Internet 
of Things (IoT) applications. While traditional machine learning and deep 
learning techniques have achieved notable success, their dependence on 
labeled data, high computational cost, and lack of robustness in edge 
environments limit their scalability. This study introduces D2H-AD, a 
novel anomaly detection framework grounded in Hyperdimensional Computing 
(HDC), designed to overcome these limitations through efficient, 
noise-resilient, and scalable mechanisms.

Although D2H-AD demonstrates strong performance, enhancing interpretability 
remains a priority. The current hypervector decoding procedure (Section 
III.G) provides feature-level explanations, but future work should 
integrate: (1) real-time visualization dashboards,
(2) confidence intervals for attribution scores, and
(3) causal inference frameworks.

The D2H-AD framework integrates both distance-based and density-aware 
encoding mechanisms within the HDC paradigm. Our empirical evaluations 
across five publicly available datasets and five comparative baseline 
methods, namely OCSVM, Isolation Forest, Autoencoder, HDAD, and ODHD, 
demonstrated that D2H-AD consistently outperforms existing approaches in 
terms of F1 score and ROC-AUC. These results validate the ability of 
D2H-AD to generalize across diverse data modalities and highlight its 
suitability for real-world anomaly detection scenarios, particularly in 
resource-constrained environments.

The hybrid density-distance scoring mechanism distinguishes D2H-AD from 
prior HDC anomaly detectors by capturing both local sparsity and global 
isolation through multiplicative fusion rather than reconstruction error 
or prototype distance alone. This algorithmic contribution demonstrates 
superior performance across diverse datasets while maintaining the 
computational efficiency inherent to HDC's binary operations. While we 
do not present hardware implementations in this work, the algorithmic 
properties of D2H-AD suggest strong potential for efficient edge 
deployment, which we leave as a direction for future hardware-focused 
research.

Despite these promising outcomes, several avenues remain open for future 
investigation. First, the incorporation of adaptive thresholding 
strategies, potentially using distribution-aware or entropy-based models, 
could further enhance detection performance under highly imbalanced or 
streaming data. For example, entropy-based adaptive thresholds can 
dynamically adjust decision boundaries in highly imbalanced datasets, 
ensuring that rare but critical anomalies are not overlooked 
\cite{deng2020self}. Second, exploring advanced encoding strategies such 
as task-specific embedding, frequency-aware quantization, or multimodal 
vector fusion may help D2H-AD better capture complex feature interactions.

Another important research direction is the real-time deployment of D2H-AD 
on ultra-low-power microcontrollers as part of the TinyML ecosystem. Given 
the method's low memory and computational requirements, evaluating its 
responsiveness and robustness under live streaming conditions would yield 
valuable insights for edge intelligence. Furthermore, enhancing model 
robustness against adversarial perturbations and data poisoning attacks, 
especially in safety-critical applications like healthcare and autonomous 
systems, remains a critical frontier.

Regarding scalability, D2H-AD's encoding phase grows linearly with dataset 
size, $O(mnD)$, making it suitable for moderately large datasets. The 
pairwise distance computation in the detection phase scales as $O(m^2D)$, 
though approximate nearest-neighbor methods can reduce this to $O(mkD)$. 
For streaming settings, the single-pass training-free architecture naturally 
supports incremental sample encoding; however, the global density estimation 
step remains a limitation in unbounded streaming scenarios. Sliding-window 
density estimation and online prototype updates are identified as important 
directions for future work.

Additionally, auto-tuning mechanisms for HDC hyperparameters (e.g., 
dimensionality, projection strategy, interval discretization) via 
meta-learning or reinforcement learning could improve the method's 
generalizability without manual intervention. Integrating explainability 
modules (e.g., interpretable similarity scores or prototype decoding) into 
D2H-AD may also increase transparency and user trust, particularly in 
domains where accountability is essential.

Investigating alternative randomness generation strategies, including 
hardware true random number generators (TRNGs), structured orthogonal 
matrices (e.g., Fast Hadamard Transform, circulant matrices), and 
deterministic quasi-random codes (e.g., Gold codes, Kasami sequences), 
presents another promising direction. These approaches could significantly 
reduce the computational overhead of hypervector generation from $O(nD)$ 
to $O(n \log D)$ while maintaining or improving orthogonality and 
separability properties. Such optimizations are particularly important 
for scaling to larger feature spaces and enabling real-time deployment on 
resource-constrained edge devices.

D2H-AD highlights the promise of hyperdimensional computing as a 
next-generation paradigm for scalable, interpretable, and energy-efficient 
anomaly detection. By further advancing adaptive modeling techniques, 
enabling deployment on embedded platforms, and incorporating security-aware 
learning strategies, this research trajectory holds substantial potential 
to enhance the robustness and intelligence of edge-based systems. We 
envision D2H-AD being deployable in TinyML environments such as wearable 
healthcare sensors and industrial IoT devices.
%%%%%%%%%%%%%%%%%%%%%%%%%%%%%%%%%%%%%%%%%%%%%%%%%%%%%%%%%%%%%%%%%%%%%%%%

\ifCLASSOPTIONcaptionsoff
  \newpage
\fi
\bibliographystyle{IEEEtran}
\bibliography{Ref}

%&&&&&&&&&&&&&&&&&&&&&&&&&&&&&&&&&&&&&&&&&&&&&&&&&&&&&&&&&

%&&&&&&&&&&&&&&&&&&&&&&&&&&&&&&&&&&&&&&&&&&&&&&&&&&&&&&&&&
%\newpage
\phantomsection
\setlength{\parskip}{0pt}
\vspace{-45pt}
\begin{IEEEbiography}
[{\includegraphics[width=1in,height=1.25in,clip,keepaspectratio]{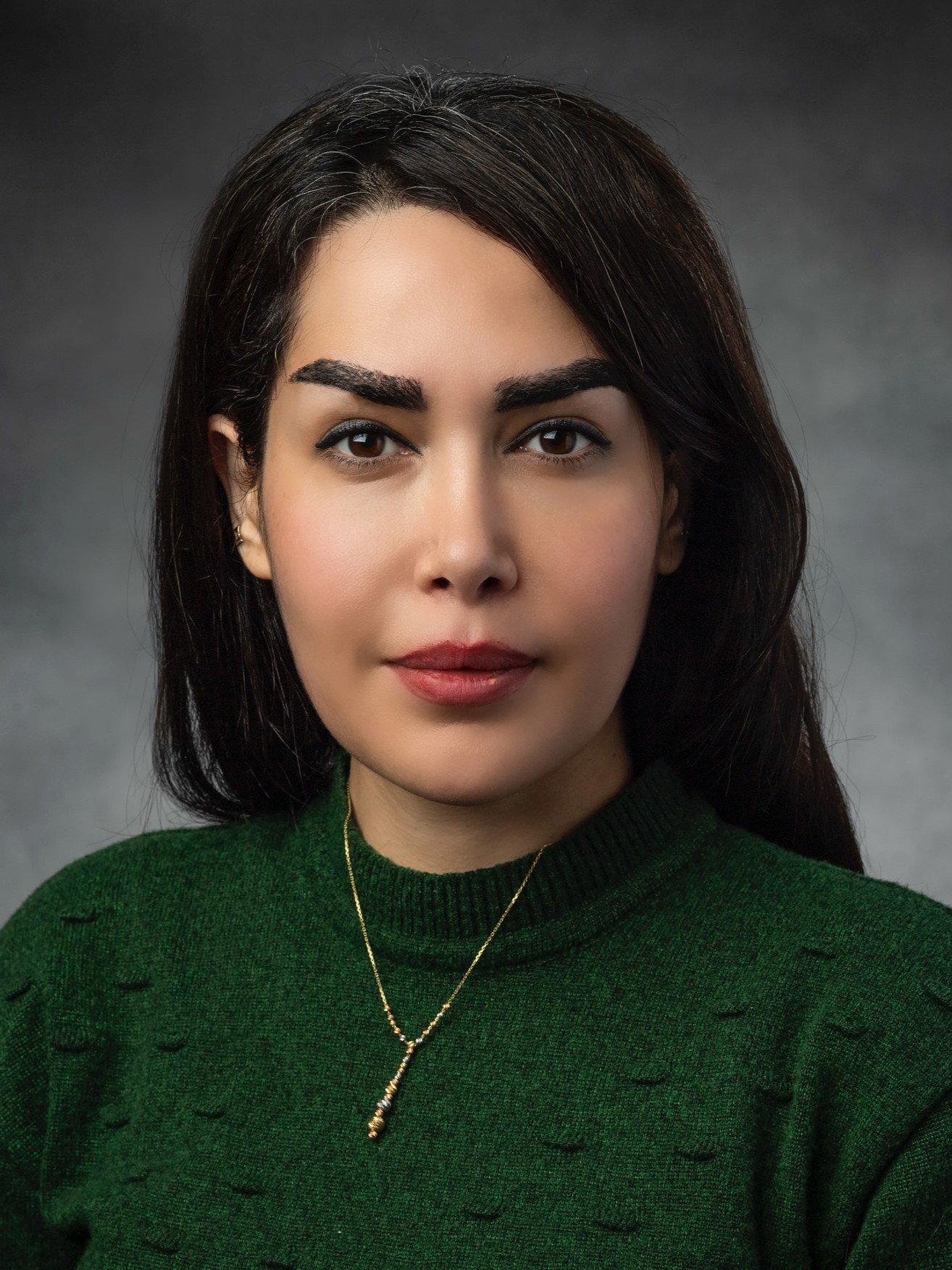}}]{Ghazal Ghajari} is a Ph.D. candidate in Computer Science at Wright State University, USA, with over a decade of experience in lecturing and applied research. Her research focuses on machine learning, hyperdimensional computing, and anomaly detection, with applications in IoT and critical infrastructure. She has authored several peer-reviewed publications and has been recognized for her excellence in scientific presentation. Beyond her publications, she actively contributes as a reviewer for leading venues in computer science, reflecting her ongoing commitment to advancing research in trustworthy AI, cybersecurity, and data-driven intelligent systems.
\end{IEEEbiography}
\vspace{-40pt} %
\begin{IEEEbiography}[{\includegraphics[width=1in,height=1.25in,clip,keepaspectratio]{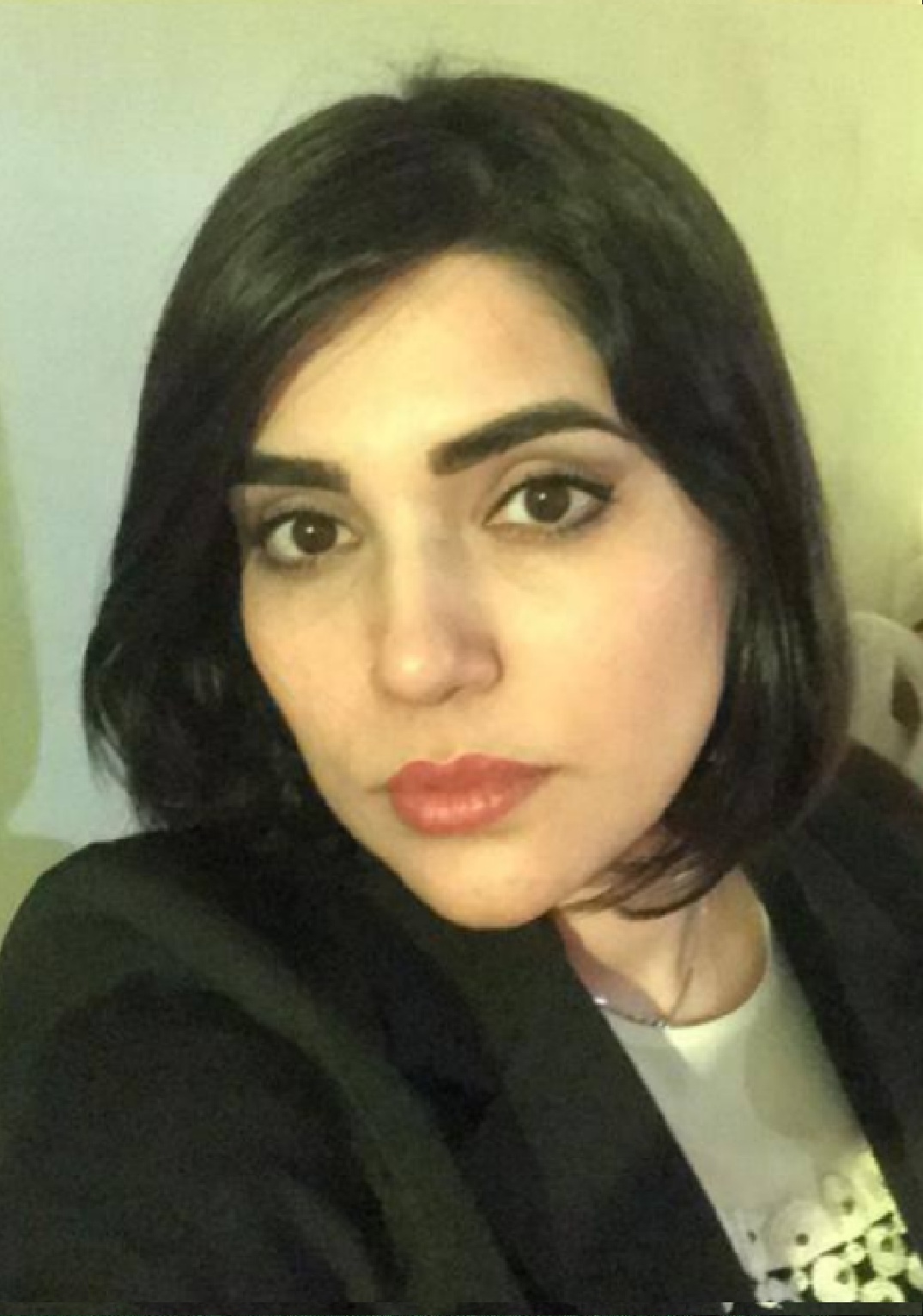}}]{Elaheh Ghajari} holds a Bachelor's degree in Computer Engineering from Islamic Azad University, Ahvaz, Iran, and has over 8 years of professional experience in programming and data analysis. Her research interests encompass machine learning, hyperdimensional computing, and anomaly detection with applications in IoT security and cybersecurity. She has co-authored multiple peer-reviewed publications accepted at prestigious IEEE conferences and serves as a reviewer for international IEEE venues. Currently working as a data analyst in the private sector, she is committed to advancing research in intelligent systems, network security, and data-driven approaches to cybersecurity challenges.
\end{IEEEbiography}
\vspace{-40pt} %
\begin{IEEEbiography}[{\includegraphics[width=1in,height=1.25in,clip,keepaspectratio]{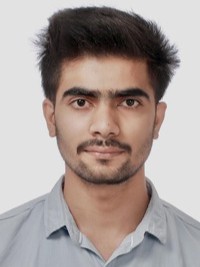}}]{Ashutosh Ghimire} is a Ph.D. candidate and Graduate Research Assistant in the Department of Computer Science and Engineering at Wright State University. He holds an M.Sc. in Computer Science from Wright State University and a B.Eng. in Computer Engineering from Tribhuvan University, Nepal. He previously worked as a Research Associate I, contributing to research in AI, software security, and hardware-based security. His interests include AI for software security, secure and privacy-aware AI hardware, side-channel analysis, trustworthy and explainable AI, adversarially resilient AI, and applications in drug discovery and signal processing. He has co-authored several publications in prominent journals and conferences, including IEEE ISVLSI and MWSCAS.\end{IEEEbiography}
\vspace{-40pt} %
\begin{IEEEbiography}[{\includegraphics[width=1in,height=1.25in,clip,keepaspectratio]{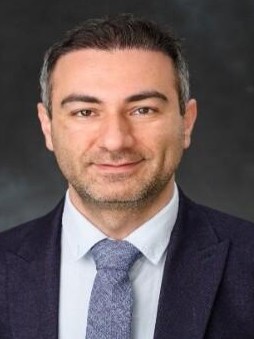}}]{Saeid Ataei} is studying Structural and Systems Engineering, specializing in complex systems and optimization. He is particularly interested in using deep learning and computer vision to enhance structural health monitoring and data analytics. With a strong background in networks and systems, he engages in projects that address real-world challenges and embraces interdisciplinary approaches to improve systems.
\end{IEEEbiography}
\vspace{-40pt} %
\begin{IEEEbiography}[{\includegraphics[width=1in,height=1.25in,clip,keepaspectratio]{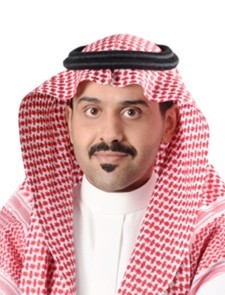}}]{Faris Alsulami}
(Member, IEEE) is an Assistant Professor with the Department of Computer and Network Engineering, College of Computer Science and Engineering, University of Jeddah, Jeddah, Saudi Arabia. He received the M.S. degree in Electrical Engineering from the University of Toledo, Toledo, OH, USA, in 2016, and the Ph.D. degree in Engineering from the University of Toledo, Toledo, OH, USA, in 2022. His research interests include hardware-assisted security for trusted embedded systems, cybersecurity, FPGA security, reverse engineering, side-channel attacks, and AI applications in software and hardware security.
\end{IEEEbiography}
\vspace{-40pt} %
\begin{IEEEbiography}[{\includegraphics[width=1in,height=1.25in,clip,keepaspectratio]{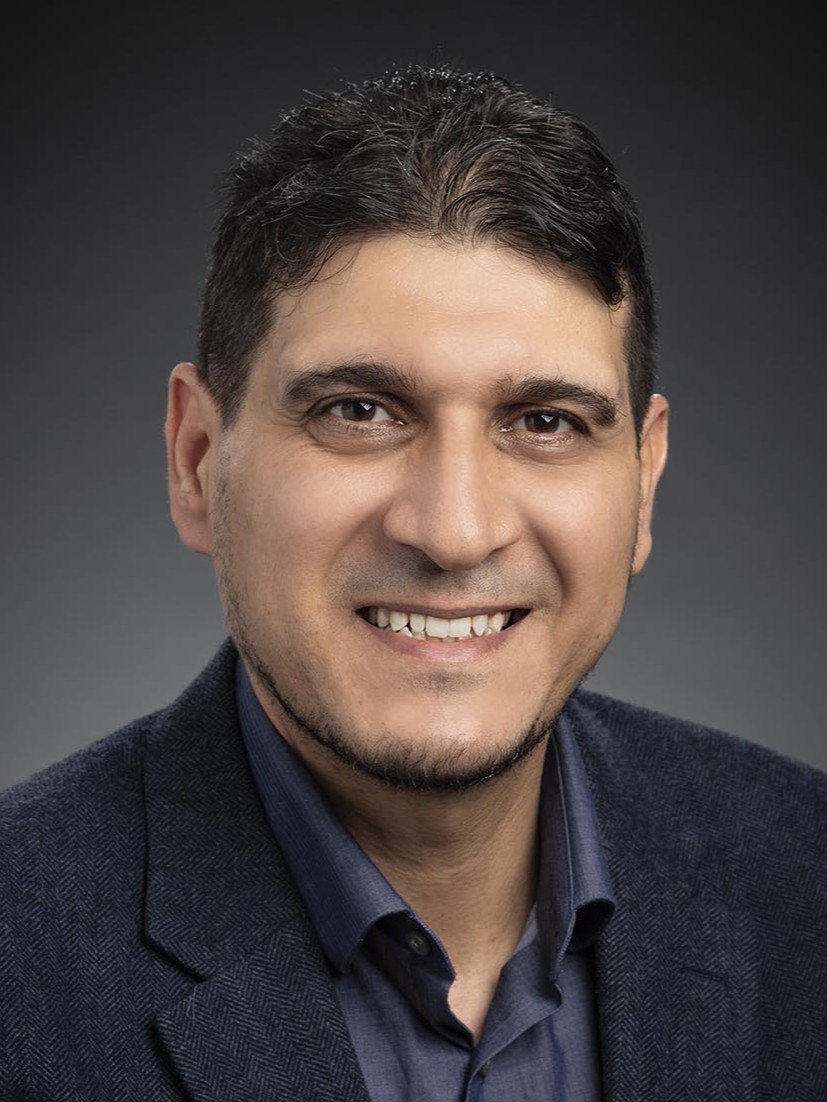}}]{Fathi Amsaad} is an Assistant Professor in the Department of Computer Science and Engineering at Wright State University, USA, with joint appointment in Biomedical, Industrial and Human Factors Engineering. He holds a PhD in Engineering from the University of Toledo and is a Senior Member of IEEE. His research focuses on hardware security, embedded system security, IoT security, TinyML, and trustworthy embedded AI. Dr. Amsaad has secured significant research funding from federal agencies including NSF, AFRL, and NSA, and has published extensively in peer-reviewed journals and conferences. He actively mentors graduate students and serves the academic community through editorial roles and conference committees in the hardware security and cybersecurity domains.
\end{IEEEbiography}

\EOD

\end{document}